\pdfoutput=1
\documentclass[12pt]{article}
\usepackage[margin=1in]{geometry}

\usepackage{times}
\usepackage{float}
\usepackage{placeins}
\usepackage{latexsym}
\usepackage{microtype}
\usepackage{hyperref}
\usepackage{datetime2}
\usepackage[most]{tcolorbox}
\usepackage[linesnumbered,ruled,vlined]{algorithm2e}
\usepackage{amsmath,amssymb}
\usepackage{subcaption}
\usepackage{caption}
\usepackage[T1]{fontenc}
\usepackage[absolute,overlay]{textpos}
\usepackage{booktabs}
\usepackage{tikz}
\usetikzlibrary{matrix,decorations.pathreplacing,calc}
\usetikzlibrary{arrows.meta}

\usepackage[utf8]{inputenc}
\usepackage{authblk}
\usepackage{longtable}
\usepackage{hyphenat}
\usepackage{tabularx}
\usepackage{booktabs}
\usepackage{subcaption}
\usepackage{csvsimple}
\usepackage{wasysym}
\usepackage[table]{xcolor}
\usepackage{siunitx}

\definecolor{HighlightA}{RGB}{200,225,255} % darker blue
\definecolor{HighlightB}{RGB}{200,245,200} % darker green
\definecolor{HighlightC}{RGB}{255,225,190} % darker orange

\usepackage[backend=biber,style=authoryear]{biblatex}
\usepackage{inconsolata}

\usepackage{graphicx}

\author[1,*]{Christo Mathew}
\author[2,*]{Wentian Wang}
\author[1]{Jacob Feldman}
\author[1]{Lazaros K. Gallos}
\author[1,3]{Paul B. Kantor}
\author[1]{Vladimir Menkov}
\author[1]{Hao Wang}

\affil[1]{Rutgers University, New Brunswick}
\affil[2]{University of Southern California}
\affil[3]{Paul B. Kantor, Consultant}
\affil[*]{These authors contributed equally  to this work}

 \title{Toward a Metrology for Artificial Intelligence: Hidden-Rule Environments and Reinforcement Learning }

\begin{document}
% \begin{center}
%     {\LARGE \bfseries Reinforcement Learning for Hidden-Rule Environments \par}
%     % \vspace{1em}
% \end{center}
% \begin{center}
%     {\centering\LARGE \bfseries Reinforcement Learning for Hidden-Rule Environments \par}
%     \vspace{1em}
% \end{center}

\begin{textblock*}{7cm}(12cm,1cm)
\raggedleft
\small
\textbf{TECHNICAL REPORT}\\
September, 2025
\end{textblock*}

\maketitle

\begin{abstract}
We investigate reinforcement learning in the Game Of Hidden Rules (GOHR) environment, a complex puzzle in which an agent must infer and execute hidden rules to clear a 6$\times$6 board by placing game pieces into buckets. We explore two state representation strategies, namely Feature-Centric (FC) and Object-Centric (OC), and employ a Transformer-based Advantage Actor-Critic (A2C) algorithm for training. The agent has access only to partial observations and must simultaneously infer the governing rule and learn the optimal policy through experience. We evaluate our models across multiple rule-based and trial-list-based experimental setups, analyzing transfer effects and the impact of representation on learning efficiency. 
\end{abstract}

% \twocolumn
\section{Introduction}

\begin{quote}
``When you can measure what you are speaking about, and express it in numbers, you know something about it; but when you cannot measure it, when you cannot express it in numbers, your knowledge is of a meagre and unsatisfactory kind: it may be the beginning of knowledge, but you have scarcely, in your thoughts, advanced to the stage of science, whatever the matter may be.'' [Lord Kelvin] \cite{thomson_electrical_1889}
\end{quote}

The new technology of Artificial Intelligence is promising to affect human society as profoundly as did the invention of the steam engine three centuries ago. Like the new technology of generative Large Language Models (LLMs), {(\cite{achiam2023gpt, comanici2025gemini, grattafiori2024llama})}
% \hao{cite some LLM papers here, like GPT-3/4 or Gemini tech report}\orange{done}
the steam engine seemed able to do the work of scores of people and do it much more quickly. And, like the steam engine, the LLMs arrive without an underlying science that would let us understand how it manages to do what it does. For the steam engine, the needed science – thermodynamics, was invented only about a century later, leading to the laws that explain the limits that Nature places on machines that convert heat to work. 

The new machines convert enormous quantities of text and other data to ``new utterances'' in response to ``prompts.'' And, as in the early 18th century, we do not understand the processes well enough to know what their limits are. Can they create genuinely new knowledge (which would be akin to a perpetual motion machine)?  Discover hidden relations? Create new languages and notations?  Or are they, like those earlier engines, constrained by some not yet discovered laws, so that they will always require our guidance and insights, as they and we advance science and technology? 

It seems likely that in another century the answers to these questions will be known – to humans, or to our symbiont or mechanical successors.  
% \hao{maybe need to have consistent styles for "–" this is different from the "––" in the abstract? These are usually considered signs of GPT-generated text these days; so we better be careful and consistent.}\orange{done, changed '--' in abstract}
For the time being, can we find a way to begin probing the limits and capabilities of these new ``thought engines?'' 

We propose that the methods of Psychology provide a very useful initial probe, towards the development of what may very well become a new science.

Psychology has, from its pre-history, and throughout its development and expansion since the nineteenth century faced the problem of understanding the workings of something that cannot be opened or examined, and which, until very recently, could only be studied physically in anomalous situations of injury or death.  Unlike the physical sciences, the field of Psychology has developed  methods for studying ``human intelligence'' (HI) by observing the behavior of (representative) humans in controlled (laboratory) situations. To do this, it has developed a host of instruments that probe the workings of the brain and mind by observing how they respond to specific stimuli and diverse challenges. In particular, there are many instruments that measure one or another aspect of ``intelligence'' – typically by asking humans (or, in some situations, animals) to solve problems. The speed and accuracy of such solutions provide fiducial points for defining various aspects of intelligence. We propose that extending these methods is a way to better understand the new machine intelligence(s) and to lay foundations for effective collaboration with them.

Let us posit that intelligence includes, at a minimum these ``traits'' : trainability, memory, discovery, and invention. We already know that AI memory is more accurate than HI. However, for HI neither laboratory methods nor introspection can identify how a given stimulus invokes a specific recollection.  Human memory seems to exploit associations and pathways that are not yet built into AI.  Let us set memory aside. Trainability and discovery may be more accessible to laboratory exploration.  Trainability (often described as ``learning'') is central in all current AI research. Systems are trained to, for example, label pictures, predict recidivism, generate job application letters, etc. We cannot ask these systems to explain what they know and so the nature of this training remains obscure.  

Generally, researchers do not in fact know an established ground truth for what it is that has been  learned, when, e.g., images are correctly labeled. That is, while we may know the correct label for each image, we do not know ``what it is about the image that makes the label correct.''  To better study trainability we would prefer a laboratory framework in which one can know whether the training has ``captured the essence'' of what is being taught.

One way to approach this is to ask the experimental subjects (in this case, AI engines of some kind) to discover a hidden rule governing correct actions, conditioned upon a situation. Canonical examples have been a staple of tests for intelligence and aptitude for many decades. One example is Ravens Progressive Matrices \cite{carpenter_what_1990}, used in many situations to screen human workers for a relevant aspect of intelligence. These typically present a number of example cases and ask the subject to choose the correct ``corresponding case'' from a small set of options. They are understood to represent relations or progressions among the three elements of an example. The subject is then challenged to repeat, extend or translate to the incomplete case. We conclude that a subject who chooses the correct answer is ``more likely'' (in a strict Bayesian sense) to have understood the rule than to have not. If there is only one correct choice among three, each such correct choice provides a factor of three in favor of ``the subject understands the rule'' against ``the subject is choosing randomly.''  To the extent that multiple examples test for ``the same concept'' correct answers quickly pile up persuasive evidence that the subject ``understands the underlying concept.'' 

Tests of this sort can be manipulated to meet Lord Kelvin’s criterion. For example, for a given specific concept (see the CMU group for a proposed classification of the concepts underlying RPM \cite{carpenter_what_1990}) one may determine, with an adequate sample of humans, the median number of training examples required to be able to solve ``similar problems'' with some specified low error rate, say five or ten percent.  The spread of those data around the median tells us something important about how ``universal'' that measured level of difficulty is. So, with some careful effort, we may be able to provide a ratio scale metric for the difficulty of learning that specific concept, in that specific framework for learning it. 

Other, more complex tests of the same kind have been developed, such as the ARC system \cite{chollet_measure_2019},\cite{moskvichev_conceptarc_2023} and have been used in a study of the kind we propose here, comparing GLLMs with human children \cite{opielka_large_2024}  A study with several kinds of abstract reasoning tests, and several kinds of AI systems has been done by \cite{gendron_large_2023}.  In this framework, the researchers seek to organize the tests into classes and to understand how the performance of various AIs differs across those classes of cognitive tasks. 

In the present note, we undertake to look carefully at the performance of several kinds of AI, on a class of well defined tasks. We ask whether the data begin to reveal the ``relative difficulty'' of these learning tasks, and whether that difficulty can be related to apparent characteristics of the tasks.  As this work progresses, it can advance  towards a set of fiducial benchmark tasks that will support development of a  new science which clarifies the relations among these diverse intelligences of contemporary AI systems.

We suggest  that one way to build for such a  new science is to concentrate on tasks that require either a human or a machine to ``discover a hidden rule.''  In order to make the task interactive, the rule should constrain a set of allowed moves, and the player, whether human or machine, should learn by trial and error. Such an interaction is not unlike taking a multiple-choice test. However, these instruments should support presenting ``the same question over and over,'' but in such a way that the answer cannot be found simply by copying what one has done before. 

One way to do this is to formulate a conceptual rule – such as ``square icons must be placed in the northwest corner of the field.''  In multiple plays of the rule, the square icons are not always shown in the same position. Thus, the learner must ``understand'' that the shape determines the allowed placement, and that position does not matter.  Rules can be made more complex, so that, for example, where a square is to be placed may depend on the quadrant of the field in which it sits. Rules can become more complex if, for example, there is one such rule for squares and another for triangles.

%\rc{Chirsto - I would like to put this next paragraph into a text box. But could not figure out how to. }

\begin{tcolorbox}[colback=gray!10, colframe=black, boxrule=0.5pt, arc=4pt]
\label{box:quote_thomson}
I may illustrate by a case in which the first step has not been taken. The hardness of different solids, as precious stones and metals, is reckoned by a merely comparative test. Diamond cuts ruby, ruby cuts quartz, quartz, I believe, cuts glass-hard steel, and glass-hard steel cuts glass; hence diamond is reckoned harder than ruby; ruby, than quartz; quartz, than glass-hard steel; and glass-hard steel, than glass: but we have no numerical measure of the hardness of  these, or of any other solids.
\flushright \cite{thomson_electrical_1889}, p.74  
\end{tcolorbox}

% exact quote: "I may illustrate by a case in which this first step has not been taken. The hardness of different solids, as precious stones and metals, is reckoned by a merely comparative test.Diamond cuts ruby, ruby cuts quartz, quartz, I believe, cuts glass-hard steel, and glass-hard steel cuts glass; hence diamond is reckoned harder than ruby; ruby, than quartz; quartz, than glass-hard steel; and glass-hard steel, than glass: but we have no numerical measure of the hardness of these, or of any other solids.  "
 % \rc{ \newline| \newline| \newline }

Using such tasks to tease out some dimensions of intelligence will be challenging. For this example framework: ``hidden rules to be discovered,'' we see that the rules may exhibit many different kinds of complexity. Therefore, we cannot expect that organizing and measuring them will  be one-dimensional in the way that hardness of materials was seen to be (See Box~\ref{box:quote_thomson}). %\rc{this wants to ref to the text box} 

Here are some very preliminary thoughts about how one might map out the ``geometry of difficulty'' in a space of hidden rules. Let’s suppose the rules are simple static rules that map shape to action. One might process a sample of different ways of describing each rule in natural language, to see whether similarity of description corresponds to similarity of difficulty. For an opposite approach one might note that the relation between two such rules can be described by a mapping of the actions, and ask whether the mapping is some kind of permutation, for which there are several known metrics, or more complex.  With research, it may be possible to establish a set of fiducial rules, which serve as surveyor's fixed points in the space of rule difficulty. 
 If so, one can use such rules to measure the capabilities of artificial intelligences, and to do so in a way that is directly applicable to a measurement of human intelligences.

Reinforcement learning (RL) has achieved remarkable progress in sequential decision-making domains, ranging from game-playing to robotics. However, many real-world tasks are governed by hidden structures or rules that are not directly observable, requiring agents to infer these underlying logics while simultaneously optimizing their actions. The Game Of Hidden Rules (GOHR) environment exemplifies such challenges: here, an agent is presented with a $6\times6$ board populated by pieces of various shapes and colors, and must deduce a hidden rule to clear the board by placing pieces into buckets.

In this work, we address the challenge of learning and generalization in the GOHR environment using RL. The agent receives only partial observations and no prior knowledge of the active rule, necessitating both effective exploration and robust representation learning. To meet these demands, we investigate two state encoding strategies: Feature-Centric (FC), which encodes global board features, and Object-Centric (OC), which encodes the properties of individual objects. In particular, FC represents actions by specifying an object's location and the bucket to which it is dropped. In the OC representation, the action is specified by naming the object and the bucket. These representations are used as inputs to a Transformer-based Advantage Actor-Critic (A2C) algorithm, \cite{mnih2016asynchronous} %\rc{this needs a citation }
allowing the agent to learn policies in an end-to-end manner. For complete documentation on the GOHR as it was originally developed by Kantor, Menkov and colleagues and the version used in this study, see \cite{menkov_rule_2022,menkov_captive_nodate,menkov_rule_nodate,menkov_setting_nodate}.

We conduct two types of experiments: (1) evaluating the agent’s ability to learn and perform different rules when trained on each rule independently, and (2) investigating transfer effects and generalization by training the agent on combinations or sequences of rules.

Our study provides new insights into the impact of state representation and training protocols in RL environments with hidden structure, and establishes GOHR as a valuable testbed for research on generalization, compositionality
of concepts, and rule inference in reinforcement learning.

\section{GOHR Environment}

The \textbf{Game Of Hidden Rules (GOHR)} is the primary environment used in our experiments. The GOHR was developed by a team of researchers at the University of Wisconsin, Madison (UWM) and is described in \cite{pulick_game_2022,bier_can_2019,bier_gohr_nodate,pulick_comparing_2024}. GOHR involves  a $6 \times 6$ board and four buckets. At the start of each game, $n(=9)$ randomly selected pieces—each with 
%distinct 
specified shape and color—are placed on the board, and a hidden rule is 
%randomly chosen.
specified. The player's objective is to clear the board by placing all pieces into the buckets, following the 
%underlying
hidden  rule, which is to be discovered during game play. Each episode terminates when all pieces have been successfully placed in any of the buckets allowed by the hidden rule. The representation of the game board is shown in Table~\ref{tab:board_representation}. In GOHR, we have two important types of entities, \emph{rules} and \emph{pieces}:
% A sample board in the web interface of CGS is shown in Fig.~\ref{fig:gohr_board} 
\begin{itemize}
    \item \textbf{Rules} define the constraints governing how pieces may be placed into buckets.

    \item \textbf{Pieces} are the game objects, each characterized by shape and color, which are randomly distributed on the board at the beginning of each episode.
\end{itemize}

\begin{table}[htb]
\centering
\setlength{\tabcolsep}{9pt}%
\renewcommand{\arraystretch}{1.15}%
\begin{tabular}{|c|*{6}{c}|c|}
\hline
\textbf{Bucket 0} & \textbf{Col1} & \textbf{Col2} & \textbf{Col3} & \textbf{Col4} & \textbf{Col5} & \textbf{Col6} & \textbf{Bucket 1} \\
(7,0) & \multicolumn{6}{c|}{} & (7,7) \\
\hline
\textbf{Row6} & 31 & 32 & 33 & 34 & 35 & 36 & \\
\textbf{Row5} & 25 & 26 & 27 & 28 & 29 & 30 & \\
\textbf{Row4} & 19 & 20 & 21 & 22 & 23 & 24 & \\
\textbf{Row3} & 13 & 14 & 15 & 16 & 17 & 18 & \\
\textbf{Row2} & 7  & 8  & 9  & 10 & 11 & 12 & \\
\textbf{Row1} & 1  & 2  & 3  & 4  & 5  & 6  & \\
\hline
\textbf{Bucket 3} & \multicolumn{6}{c|}{} & \textbf{Bucket 2} \\
(0,0) & \multicolumn{6}{c|}{} & (0,7) \\
\hline
\end{tabular}
\caption{Representation of the 6$\times$6 board with four corner buckets. 
Each numbered cell corresponds to a \textbf{position index} (1–36), starting from the bottom-left (Col1, Row1) to the top-right (Col6, Row6). 
Coordinates are given as $(x,y)$ pairs, where $x$ increases from left (0) to right (7) and $y$ increases from bottom (0) to top (7). 
Buckets are placed at the four corners: 
Bucket~0 at $(7,0)$, Bucket~1 at $(7,7)$, Bucket~2 at $(0,7)$, and Bucket~3 at $(0,0)$. 
Thus, the numbering (1–36) provides a linear position index for learning tasks, while coordinates encode the true spatial layout. }
\label{tab:board_representation}
\end{table}

% \begin{figure}[htbp]
%     \centering
%     \includegraphics[width=0.8\textwidth]{board_webinterface.png}
%     \caption{The web interface of the board. the right side shows the board with 9 pieces. 2 pieces already assigned to bucket is shown with green \checkmark. And 4  buckets are in the corner. The left side shows the response of 3 moves first 2 moves were successful, shown as \smiley  and $3^{rd}$ move was unsuccessful shown as 
%     \frownie}.
%     \label{fig:gohr_board}
% \end{figure}
% \subsection{State Space}

At each timestep, the environment provides an observation comprising the current board state and a set of status codes.

\subsection{Board State}
% \hao{why use subsubsection without subsection? Same issue for all following subsubsections.}\orange{done, corrected section numbering}
The board state is represented as a list of all pieces on the board, along with their coordinates $(x, y)$. Each entry specifies the features (shape, color) and position of a piece.

\subsection{Status Codes}
\label{sec:response-codes}

The status codes consist of three integer values: \textbf{response\_code}, \textbf{finish\_code}, and \textbf{move\_count}. Specifically:
\begin{itemize}
    \item \textbf{response\_code:} Indicates whether the previous action was accepted or rejected. {Possible values are 0 (ACCEPT, a successful move), 4 (DENY), and 7 (IMMOVABLE).} 
    \item \textbf{finish\_code:} Represents the termination status of the game (e.g., ongoing, completed, failed).
    \item \textbf{move\_count:} Tracks the total number of moves attempted so far.
\end{itemize}

{Our RL agent relies on these status codes to obtain feedback from the environment. In particular, \textbf{response\_code} values are logged as \textit{A}, \textit{D}, and \textit{I} for codes 0, 4, and 7, respectively, and are used when computing the evaluation metric $m\_star$. The \textbf{finish\_code} is used to determine whether an episode is successfully completed within the allowed number of moves. The \textbf{move\_count} is currently ignored by the agent during training.}

% \hao{maybe add more details on why this is relevant to RL or simply say that our RL does not use these information yet}\orange{done, please let me know if anything further is required}

Further details about the GOHR environment can be found at: \url{http://action.rutgers.edu/w2020/captive.html#cond}

\subsection{Action Space}

{At each timestep, the agent chooses an action corresponding to assigning a piece to one of the four buckets. The exact structure of the action space depends on the state representation.} 
% \hao{give some examples here?}\orange{done, added examples in the following paragraph.}

{\paragraph{Feature-Centric (FC) representation.}  
In the FC setting, actions are indexed by board position. Each of the $6 \times 6 = 36$ positions has 4 possible actions, one for each bucket. Thus the FC action space is
\[
A_{\text{FC}} = \{\, a \in \mathbb{Z} \mid 0 \leq a \leq 143 \,\}.
\]
An action $a$ corresponds to “move the piece at position $(x,y)$ into bucket $b \in \{0,1,2,3\}$.”  {These actions are encoded as one-hot vectors of length 144, with a 1 at the index corresponding to the chosen action and 0 elsewhere. }
For example, actions $0,1,2,3$ represent assigning the piece at position $1$ (first cell) to buckets $0,1,2,3$, respectively; actions $4,5,6,7$ correspond to assigning the piece at position $2$ to buckets $0,1,2,3$; and so on.} 
% \hao{Also mention that we use one-hot encoding for $A_{FC}$, so it is actually a 144D vector. Same for OC below.}\orange{[done, included explanation of onehot encoding in FC and OC.]}

{\paragraph{Object-Centric (OC) representation.}  
If $n$ objects are present on the board, the action space consists of $n \times 4$ actions, one for each possible object–bucket assignment. Actions for each object are encoded as one-hot vectors of length 4, indicating the chosen bucket (see Fig.~\ref{fig:stateoc}). For example, when $n=9$, the action space is
\[
A_{\text{OC}} = \{\, a \in \mathbb{Z} \mid 0 \leq a \leq 35 \,\}.
\]
In this case, actions $0,1,2,3$ correspond to assigning object $1$ to buckets $0,1,2,3$, and so on . 
%respectively; actions $4,5,6,7$ correspond to assigning object $2$ to buckets $0,1,2,3$; and so forth.}

\subsection{Reward Function}

The reward structure is defined as follows: each successful move (i.e., a valid placement of a piece into a bucket which is {indicated by response code of 0}) yields a reward of $0$, while each invalid move {(indicated by response codes 4 and 7)} results in a reward of $-1$. 

\begin{figure}[t]
  \centering
  \includegraphics[width=0.96\linewidth]{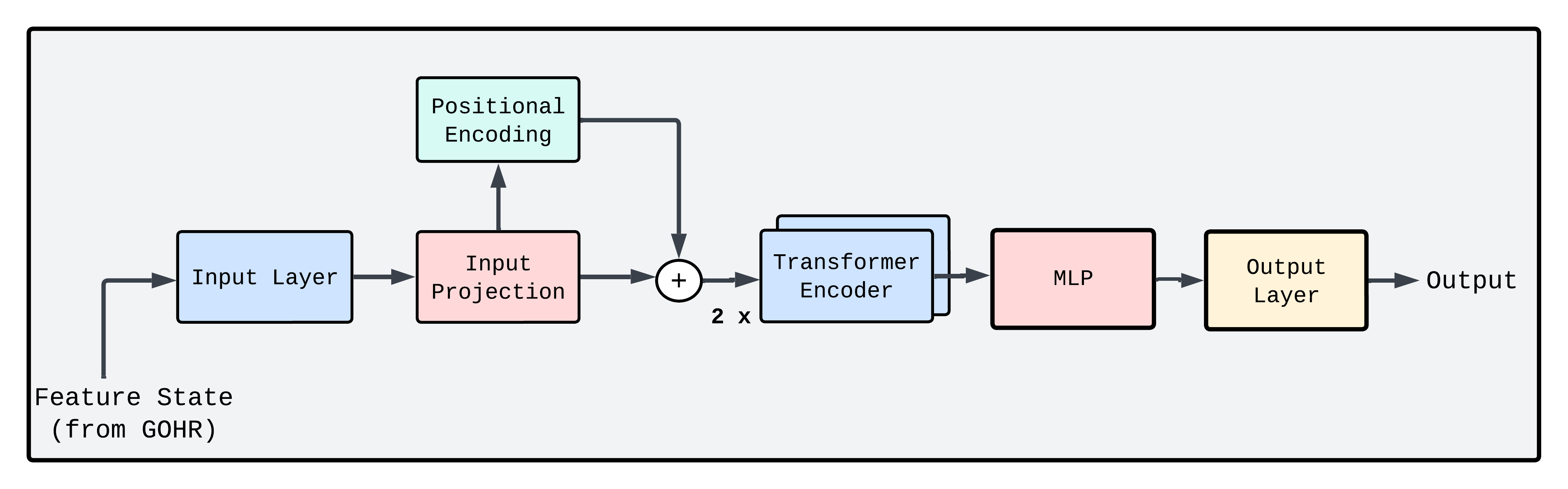} \hfill
  \caption {Transformer architecture used for policy network and critic network. }
\end{figure}
\

\section{Methodology}

\subsection{State Representation and Feature Engineering}

To facilitate effective model training, the raw states provided by the GOHR game server are transformed using two principal strategies: \textbf{Feature-Centric Representation (FC)} and \textbf{Object-Centric Representation (OC)}.

\begin{figure}[ht]
  \centering
  % --- statefc_fig.tex (no \documentclass, no \usepackage) ---
\begingroup
% Parameters
\def\GRIDSIZE{6}
\def\LAYERS{8}
\def\CELLSZ{0.6}
\def\DX{0.5}
\def\DY{0.3}

% Boolean used to mark whether the current cell is a 1
\newif\ifcellone

% Draw a layer with multiple ones
% Usage:
%   \DrawMultiLayer{sx}{sy}{5/1}                 % one position
%   \DrawMultiLayer{sx}{sy}{2/1,5/3,3/6}         % multiple positions
\newcommand{\DrawMultiLayer}[3]{%
  \begin{scope}[shift={(#1,#2)}]
    \fill[white] (0,0) rectangle (\GRIDSIZE*\CELLSZ,\GRIDSIZE*\CELLSZ);
    \foreach \a in {0,...,\GRIDSIZE}{
      \draw[gray!40, thin] (\a*\CELLSZ,0) -- (\a*\CELLSZ,\GRIDSIZE*\CELLSZ);
      \draw[gray!40, thin] (0,\a*\CELLSZ) -- (\GRIDSIZE*\CELLSZ,\a*\CELLSZ);
    }
    \draw[black,thick] (0,0) rectangle (\GRIDSIZE*\CELLSZ,\GRIDSIZE*\CELLSZ);

    \foreach \r in {1,...,\GRIDSIZE}{
      \foreach \c in {1,...,\GRIDSIZE}{
        \pgfmathsetmacro{\x}{(\c-1)*\CELLSZ}
        \pgfmathsetmacro{\y}{(\GRIDSIZE-\r)*\CELLSZ}

        % Is this cell a 1?
        \global\cellonefalse
        \foreach \rr/\cc in {#3}{%
          \ifnum\r=\rr\relax
            \ifnum\c=\cc\relax
              \global\cellonetrue
            \fi
          \fi
        }

        \ifcellone
          \fill[blue!25] (\x,\y) rectangle ++(\CELLSZ,\CELLSZ);
          \node[font=\bfseries] at (\x+0.5*\CELLSZ,\y+0.5*\CELLSZ) {1};
        \else
          \node[gray, font=\small] at (\x+0.5*\CELLSZ,\y+0.5*\CELLSZ) {0};
        \fi
      }
    }
  \end{scope}
}

\begin{tikzpicture}
  % Layer 1
  \DrawMultiLayer{0}{0}{5/1, 1/3}
  \node[font=\scriptsize, anchor=west] at (3.9, 3.6) {$f_1$};

  % Layer 2
  \DrawMultiLayer{0.5}{-0.3}{1/1,1/6}
  \node[font=\scriptsize, anchor=west] at (4.4, 3.3) {$f_2$};

  % Layer 3
  \DrawMultiLayer{1.0}{-0.6}{3/1,4/3}
  \node[font=\scriptsize, anchor=west] at (4.9, 3.0) {$f_3$};

  % Layer 4
  \DrawMultiLayer{1.5}{-0.9}{6/1,1/4}
  \node[font=\scriptsize, anchor=west] at (5.4, 2.7) {$f_4$};

  % Layer 5
  \DrawMultiLayer{2.0}{-1.2}{6/1,1/3,2/5}
  \node[font=\scriptsize, anchor=west] at (5.9, 2.4) {$f_5$};

  % Layer 6
  \DrawMultiLayer{2.5}{-1.5}{1/1,1/6}
  \node[font=\scriptsize, anchor=west] at (6.4, 2.1) {$f_6$};

  % Layer 7
  \DrawMultiLayer{3.0}{-1.8}{5/1}
  \node[font=\scriptsize, anchor=west] at (6.9, 1.8) {$f_7$};

  % Layer 8
  \DrawMultiLayer{3.5}{-2.1}{3/1,1/4,2/3,5/3 }
  \node[font=\scriptsize, anchor=west] at (7.4, 1.5) {$f_8$};

  % Brace and caption (keep these if you want them part of the graphic)
  \coordinate (bracestart) at (4,3.8);
  \coordinate (braceend)   at (7.8,1.5);
  \draw[decorate,decoration={brace,amplitude=8pt,raise=5pt},thick]
    (bracestart) -- (braceend)
    node[midway,above=12pt,sloped,font=\small]{No. of features $=8$};

  \node[font=\small, text width=6cm, align=center] 
    at (.5,-1.5) {$6 \times 6 \times 8$ tensor};
\end{tikzpicture}
\endgroup
  \caption{Feature-Centric (FC) representation of a single board at a given time step. The board is encoded as a stacked one-hot tensor with 8 feature maps, each of size $6 \times 6$.
  Features $f_1$–$f_8$ correspond to square, star, circle, triangle, red, black, blue, and yellow, respectively. The example shows the placement of 9 objects, such as a red triangle at position 1 and a yellow star at position 9.}
  % \caption{\textbf{Stacked one-hot tensor representation of a single board at one time-step}. The tensor consists of 8 feature maps, each of dimension $6 \times 6$. 
  % Features $f_1$–$f_8$ correspond to square, star, circle, triangle, red, black, blue, and yellow, respectively. The example illustrates the placement of 9 objects, such as red triangle at postion 1 and yellow star at position 9.}  
  %\hao{for each position, there should be a "1" in the first 4 features and a "1" in the remaining 4 features. This figure's numbers need to revised.}\orange{done, corrected the positions and included some examples in caption.Also updated its explanation in main text(highlighted in blue)}

  % square, star, circle, triangle, red, black, blue and yellow
  \label{fig:statefc}
\end{figure}
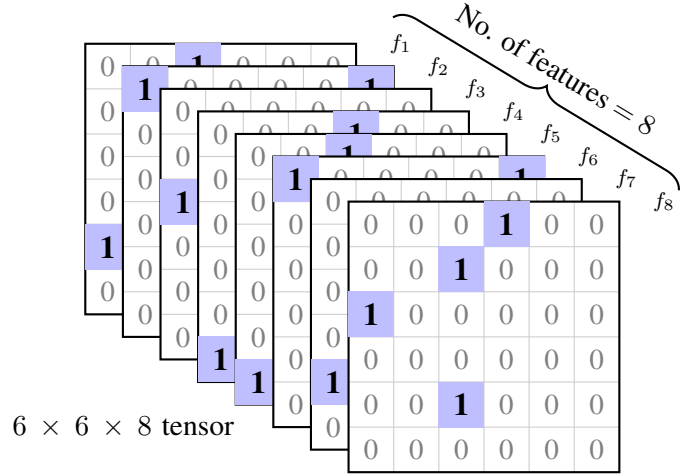

\paragraph{Feature-Centric (FC) Representation:} 
% In the FC approach, the state is encoded as a collection of feature maps over the $6 \times 6$ grid. Each feature (such as shape or color) is represented as a $36$-dimensional binary vector. For instance, the \emph{shape} attribute is encoded as a $4 \times 36$ binary matrix, where each row denotes a distinct shape and each entry indicates the presence ($1$) or absence ($0$) of that shape in a cell. The action space is similarly represented by a vector of size $144$ ($6 \times 6 \times 4$), corresponding to all possible assignments of pieces to buckets.

In the Feature-Centric (FC) representation,the state is encoded as a collection of feature maps over the $6 \times 6$ grid. Each feature type (e.g., a shape, a color) is encoded as a 36-dimensional one-hot vector, corresponding to the 36 grid positions on the board. For example, if there is a square at position 35, the ``square'' vector will have a 1 at index 35. Likewise, if there is a red object at that same position, the “red” vector will have a 1 at index 35. 
% \hao{need an example here; maybe use a 8D vector with 4 entries for color and 4 entries for shape; please use math notation here to be rigorous. you can also refer readers to section 2.1.2 below where you have more details}\orange{done, added a diagram and explanation}
Therefore,  the system does not directly encode ``red square'' as a single combined feature; instead, it provides parallel position-specific signals for “square” and for ``red.'' The model can infer that they belong to the same object because both the ``square'' and the ``red'' vectors light up at the same position. In that sense, the FC representation does give the model a way to connect color and shape features through positional alignment, but it requires the model to learn the association — there is no explicit fused ``red square'' input. 

{Fig.~\ref{fig:statefc} illustrates the full feature representation of a board at a single time step. Each of the eight $6 \times 6$ grids (denoted $f_1, f_2, \dots, f_8$) encodes one feature map, with details of the feature assignments provided in the caption.}

\begin{figure}[ht]
  \centering
  \begin{tikzpicture}[font=\ttfamily]
% The bitstrings are arranged as 5 columns in a TikZ matrix:
%   col1=Color(4), col2=Shape(4), col3=x(6), col4=y(6), col5=Action(4)
\matrix (M) [matrix of nodes,
  nodes={anchor=west, inner sep=0pt, outer sep=0pt},
  column sep=8mm, row sep=1.2ex] {
  1000 & 0100 & 100000 & 000001 & 0000 \\
  0001 & 0001 & 010000 & 000001 & 0000 \\
  1000 & 0010 & 000001 & 000010 & 0100 \\
  0010 & 1000 & 000100 & 001000 & 0000 \\
  0100 & 0001 & 001000 & 100000 & 0000 \\
  1000 & 0100 & 000010 & 010000 & 0000 \\
  0001 & 0010 & 100000 & 000100 & 0000 \\
  0010 & 0001 & 010000 & 001000 & 0000 \\
  0100 & 1000 & 000001 & 000001 & 0000 \\
};

% ---- Braces ABOVE each feature group with labels ----
\foreach \col/\label in {1/Color,2/Shape,3/$x$,4/$y$,5/Action} {
  \draw[decorate, decoration={brace, amplitude=3pt}, yshift=6pt]
    (M-1-\col.north west) -- (M-1-\col.north east)
    node[midway, yshift=8pt, font=\sffamily\footnotesize]{\label};
}

% ---- (Optional) Braces BELOW instead of above: uncomment to use ----
% \foreach \col/\label in {1/Color,2/Shape,3/$x$,4/$y$,5/Action} {
%   \draw[decorate, decoration={brace, mirror, amplitude=3pt}, yshift=-6pt]
%     (M-9-\col.south west) -- (M-9-\col.south east)
%     node[midway, yshift=-8pt, font=\sffamily\footnotesize]{\label};
% }

% ---- (Optional) subtle column backgrounds for extra clarity ----
% \foreach \col in {1,...,5} {
%   \begin{scope}[on background layer]
%     \fill[black!5, rounded corners=1pt]
%       ($(M-1-\col.north west)+(-1pt,3pt)$) rectangle
%       ($(M-9-\col.south east)+(1pt,-3pt)$);
%   \end{scope}
% }

\end{tikzpicture}
  \caption{Object-Centric (OC) representation of a single board at a given time step. The board is encoded as a one-hot tensor with 9 rows, each corresponding to one object on the board. Every object is represented as a 24-dimensional one-hot vector, partitioned into color, shape, $x$, and $y$ components. The grouping illustrates how these components form the OC model input representation.}
  % \caption{One-hot tensor representation of a single board at a time-step. The representation consists of 9 rows, each corresponding to one object on the board. Every object is encoded as a 24-dimensional one-hot vector. The grouping highlights how these 24 dimensions are partitioned into color, shape, $x$, and $y$ components. }
  \label{fig:stateoc}
\end{figure}

\paragraph{Object-Centric (OC) Representation:}
In the OC approach, each piece on the board is represented as a $20$-dimensional feature vector, with one-hot encodings for color, shape, and $(x, y)$ position. {Color and shape are encoded as one-hot vectors of dimension 4 each while $x$ and $y$ are encoded as one-hot vectors of dimension 6 each.} 
% \hao{mention how do you add up to 20} \orange{done, explained and added Fig.~\ref{fig:stateoc}}
Actions are associated with objects: for each object, a one-hot vector of length $4$ represents the possible placements into the four buckets. Therefore, for a configuration with $n$ pieces and 4 buckets, the total number of possible actions is $4\times n$.
{Consider a \textbf{red} object(piece) of \textbf{square} shape present in \textbf{column1} and \textbf{row 6} and is attempted to move to \textbf{bucket 2}, this can be represented as 1000 0100 100000 000001 0010. Fig.~\ref{fig:stateoc}} shows the complete representation of a board in a timestep.

\subsection{Input and Output Representation}

\paragraph{Input:}
At each step, the model receives as input the current board state together with the six \footnote{In some runs this is increased to eight, which can cover an entire episode for removing nine pieces} most recent successful board states and their corresponding actions. This composite input can be described as:
%\begin{multline}
\begin{equation}
    \text{Input} = \Big[\, 
    \text{CurrentState},\, 
    \text{PastState}_1,\, \text{PastState}_2,\, \ldots,\, \text{PastState}_6,\, 
    \,\Big]
\end{equation}
%\end{multline}

% \begin{multline}
%     \text{Input} = \Big[\, 
%     \underbrace{\text{CurrentBoard}}_{288},\, 
%     \underbrace{\text{PastBoard}_1,\, \text{Action}_1,\, \ldots,\, \text{PastBoard}_6,\, \text{Action}_6}_{6 \times (288 + 144)}
%     \,\Big]
% \end{multline}
% \hao{need to explain where does 288 and 144 above come from. also it seems the above is only for FC not OC? need to clarify this}

\begin{figure}[h]
      \centering
      \resizebox{\columnwidth}{!}{
        % --- complete_statefc_flow.tex (no \documentclass, no \usepackage) ---
\begingroup
% ---------- Parameters ----------
\def\GRIDSIZE{6}
\def\LAYERS{8}
\def\CELLSZ{0.4}
\def\DX{0.32}   % horizontal offset per ghost layer
\def\DY{0.24}   % vertical offset per ghost layer

% ---------- Draw a stacked 2D board (ghosts behind, white label on top) ----------
% Ghost layers extend right (+x) and down (-y) so the stack reads left->right, top->bottom.
% A tiny bleed (\EPSX,\EPSY) lets the ghosts peek from the right & bottom so it feels continuous.
\newcommand{\DrawMultiLayer}[3]{%
  \begin{scope}[shift={(#1,#2)}]
    \fill[white] (0,0) rectangle (\GRIDSIZE*\CELLSZ,\GRIDSIZE*\CELLSZ);
    \foreach \a in {0,...,\GRIDSIZE}{
      \draw[gray!40, thin] (\a*\CELLSZ,0) -- (\a*\CELLSZ,\GRIDSIZE*\CELLSZ);
      \draw[gray!40, thin] (0,\a*\CELLSZ) -- (\GRIDSIZE*\CELLSZ,\a*\CELLSZ);
    }
    \draw[black,thick] (0,0) rectangle (\GRIDSIZE*\CELLSZ,\GRIDSIZE*\CELLSZ);
    
    % Simplified - just show a few highlighted cells
    \foreach \r/\c in {#3}{
      \pgfmathsetmacro{\x}{(\c-1)*\CELLSZ}
      \pgfmathsetmacro{\y}{(\GRIDSIZE-\r)*\CELLSZ}
      \fill[blue!25] (\x,\y) rectangle ++(\CELLSZ,\CELLSZ);
    }
  \end{scope}
}

% Command to draw a simplified 3D stack
\newcommand{\DrawStack}[4]{%  {x}{y}{label}{highlight_cells}
  \begin{scope}[shift={(#1,#2)}]
    % Draw multiple layers with offset
    \foreach \i in {0,...,4}{
      \pgfmathsetmacro{\offset}{\i*0.05}
      \DrawMultiLayer{\offset}{\offset}{#4}
    }
    \node[font=\scriptsize, anchor=center] at (1.2, -0.3) {#3};
  \end{scope}
}

\begin{tikzpicture}[>=stealth]
\begin{scope}[shift={(-2,0)}]
% ================= Left side: two stacked boards =================
\DrawStack{0}{4}{state\_fc}{1/1,3/2,5/4}

\draw[->, thick] (2.8, 5.6) -- (4.1, 5.6);
\node[font=\scriptsize, above] at (3.4, 5.6) {flatten};

\draw[thick] (4.3, 5.2) rectangle (6.3, 6.0);
\node[font=\scriptsize, align=center] at (5.3, 5.6) {Flattened\\$\mathbb{R}^{288}$};

\draw[thick] (4.3, 4.2) rectangle (6.3, 5.0);
\node[font=\scriptsize, align=center] at (5.3, 4.6) {Actions\\$\mathbb{R}^{144}$};

\draw[->, thick] (6.5, 5.6) -- (7.7, 5.2);
\draw[->, thick] (6.5, 4.6) -- (7.7, 5.0);
% \node[font=\scriptsize, right] at (6.7, 4.5) {concat};

% Current board - positioned at coordinates (0,1.5) 
\DrawStack{0}{.5}{current\_board}{1/1,3/2,5/4}

\draw[->, thick] (2.8, 1.8) -- (4.1, 1.8);
\node[font=\scriptsize, above] at (3.4, 1.8) {flatten};

\draw[thick] (4.3, 1.3) rectangle (6.3, 2.3);
\node[font=\scriptsize, align=center] at (5.3, 1.8) {Flattened\\$\mathbb{R}^{288}$};

% ================= Right side: vertical trio =================
% Top: Past State
\draw[thick] (8.0, 4.7) rectangle (10.6, 5.5);
\node[font=\scriptsize, align=center] at (9.3, 5.1) {Past State\\$\mathbb{R}^{432}$};

% Middle: 6 × Past State
\draw[thick] (8.0, 3.2) rectangle (10.6, 4.0);
\node[font=\scriptsize, align=center] at (9.3, 3.6) {$6 \times$ Past State\\$\mathbb{R}^{2592}$};

% Arrow top -> middle
\draw[->, thick] (9.3, 4.7) -- (9.3, 4.05);
\node[font=\scriptsize, right] at (9.55, 4.3) {$\times 6$};

% Bottom: Concat(current, 6 × past)
\draw[thick, fill=green!5] (7.2, 1.2) rectangle (10.6, 2.4);
\node[font=\scriptsize, align=center] at (8.9, 1.8)
  {Concat$(\text{current},\, 6\times\text{past})$\\$\mathbb{R}^{2880}$};

% Inputs into bottom merge
\draw[->, thick] (6.5, 1.8) -- (7.0, 1.8);     % from flattened current
\draw[->, thick] (9.3, 3.2) -- (9.3, 2.45);    % from 6× past

% Final Input to FC (green arrow only from bottom box)
\draw[thick, fill=green!10] (11.4, 1.2) rectangle (14.2, 2.4);
\node[font=\small, align=center] at (12.8, 1.8) {Input to FC\\$\mathbb{R}^{2880}$};
\draw[->, thick, green!70!black] (10.8, 1.8) -- (11.3, 1.8);

% % Small blue dimension tags
% \node[font=\tiny, color=blue] at (6.1, 5.2) {$288$};
% \node[font=\tiny, color=blue] at (6.1, 4.2) {$144$};
% \node[font=\tiny, color=blue] at (10.3, 4.7) {$432$};
% \node[font=\tiny, color=blue] at (10.3, 3.2) {$2592$};
% \node[font=\tiny, color=blue] at (14.1, 1.5) {$2880$};

\end{scope}
% Footnote
\node[font=\scriptsize, align=left] at (0, -0.5) {
  Dimensions: $288 + 2592 = 2880$\\
  Where: $6 \times 432 = 2592$
};

\end{tikzpicture}
\endgroup
      }
      \caption{Feature construction pipeline of FC model. Top: Past state formed by concatenating the flattened FC state (\(\mathbb{R}^{288}\)) with the action encoding (\(\mathbb{R}^{144}\)), yielding \(\mathbb{R}^{432}\). Bottom: Current input\(_{\text{FC}}\) formed by concatenating the current flattened board (\(\mathbb{R}^{288}\)) with six past states (\(6\times \mathbb{R}^{432}\)), producing \(\mathbb{R}^{2880}\).}
      \label{fig:statefc-pipeline}
    \end{figure}
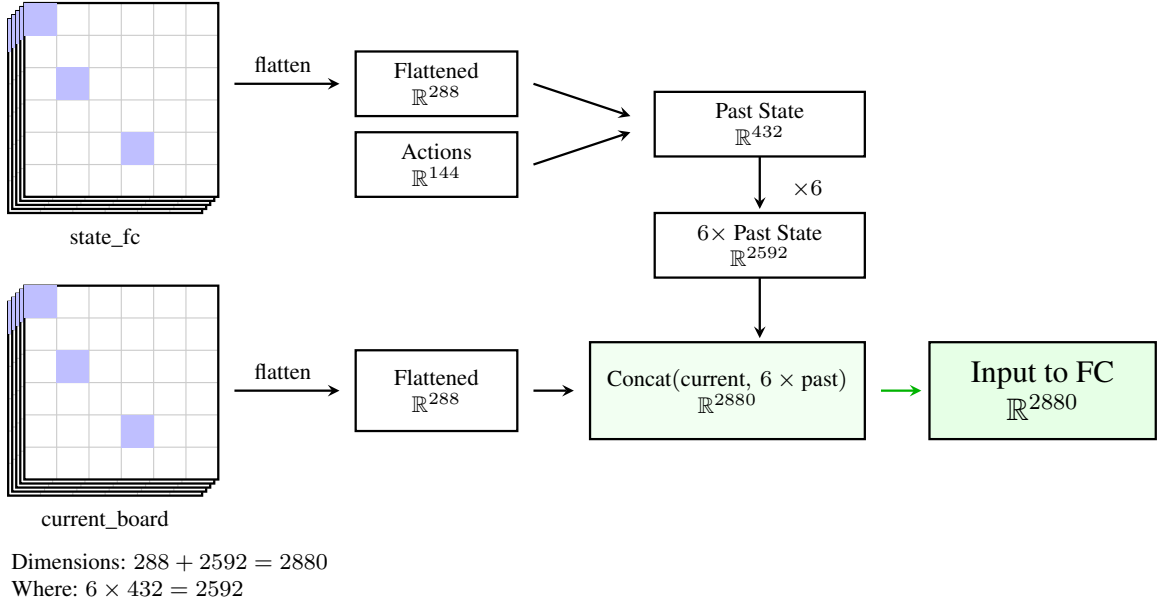
    
    \begin{figure}[h]
      \centering
      \resizebox{\columnwidth}{!}{
        \input{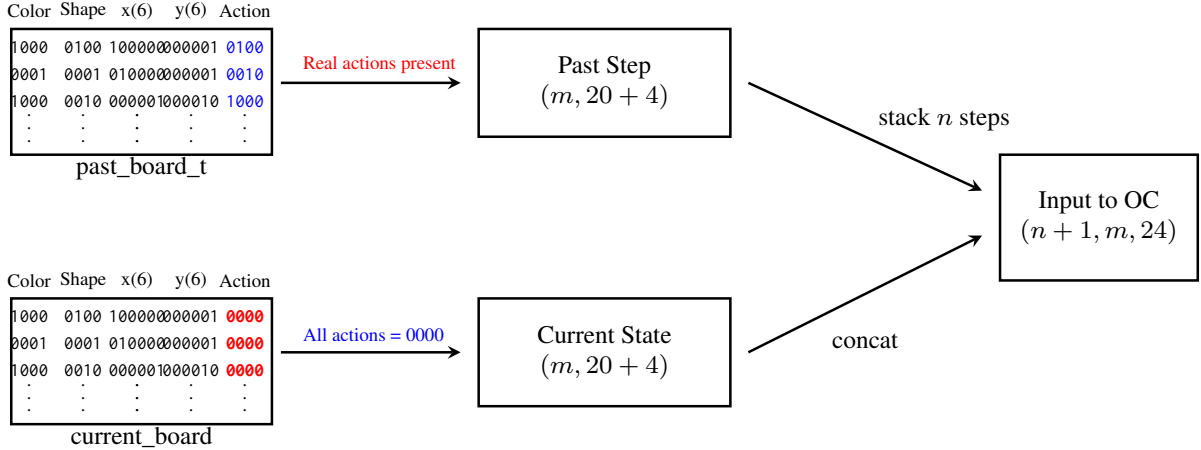}
       }
    
      \caption{Feature construction pipeline of OC model. $m$ is the number of objects, n is the number of previous board states. }
      \label{fig:stateoc-pipeline}
    \end{figure}
The input is encoded in two different ways, depending on the model architecture:
\begin{itemize}
    \item \textbf{Feature-Centric (FC) model:} The composite input is flattened into a vector of dimension $2880$, i.e., $\text{Input}_{\mathrm{FC}} \in \mathbb{R}^{2880}$. Refer Fig.~\ref{fig:statefc-pipeline}.

    \item \textbf{Object-Centric (OC) model:} The input is structured as a tensor of shape $7 \times n \times 24$, i.e., $\text{Input}_{\mathrm{OC}} \in \mathbb{R}^{7 \times n \times 24}$, where $n$ is the number of objects per board state and each object is described by a $24$-dimensional feature vector ($20$ features and $4$ for actions).
    % , with zero-padding for actions in the current state). 
    % \hao{what do you mean by "with zero-padding for actions in the current state", do you mean for pieces without action you simply use 0 for all 4 entries?}\orange{christo: for the current board 4 digits corresponding to actions are padded to zero so dimensions will be correct. }
    Refer Fig.~\ref{fig:stateoc-pipeline}.
    % \hao{Please (1) use Fig. X with upper case to refer to figures and (2) do not forget "." at the end of the sentence after referring to Figures. Check this typo throughout the paper.}\christo{done , changed to Fig.}

\end{itemize}

\begin{algorithm}[!htb]
\caption{A2C Training Algorithm for GOHR (with Entropy Regularization) 
% \hao{there is an extra ";" in line 15? also you do not need do describe $\beta$ inside the algorithm if it makes the layout messy; just describe it in the main text. Also could we use a smaller page margin so that we have more space per line for this algorithm?}\christo{has modified algorithm by referencing eqns and updating notations.}
}
\KwIn{Environment $\mathcal{E}$; policy network $\pi_\theta(a \mid s)$; critic $V_\phi(s)$; hyperparameters $(\gamma,\,n_{\text{episodes}},\,T,\,\beta)$; 
action mask function $m_t$; valid action set ${v}_t$}

Initialize $\pi_\theta$ and $V_\phi$ with random weights\;

\For{episode $= 1$ to $n_{\text{episodes}}$}{
    Reset environment $\mathcal{E}$, 
    observe initial state $s_0$, mask $m_0$, valid set ${v}_0$\;
    % \hao{you would need to have notations for mask and valid actions too. you need to say the variable name first and then the variable; e.g., use "observe initial state $s_0$".}\christo{done}\; 
    Initialize lists $S \gets [~],\ A \gets [~],\ R \gets [~],\ \log P \gets [~],\ Ent \gets [~]$\;

    \For{$t = 0$ \KwTo $T-1$}{
        Sample action $a_t \sim \pi_\theta(\cdot \mid s_t, m_t,{v}_t)$ (e.g., $\epsilon$-greedy or stochastic)\;
        Compute log-probability $\ell_t =\log \pi_\theta(a_t \mid s_t)$ and entropy $H_t = -\sum_{a} \pi_\theta(a\mid s_t) \log \pi_\theta(a\mid s_t)$\;
        Execute $a_t$ in $\mathcal{E}$, observe $r_t$,$s_{t+1}$, { ${v}_{t+1}$, and terminal flag $\texttt{done}$\;} 
        Store $s_t$, $a_t$,$r_t$, $\ell_t$ and $H_t$ in $S$, $A$, $R$, $\log P$, and $Ent$\ respectively \;
        \If{$\texttt{done}$}{
            \textbf{break}\;
        }
    }    

    Compute returns $G_t$ (Eq.~\ref{eq:returns})\;
    % \hao{write down the exact equation in the main text and refer to this equation here}\christo{done}\;
    Compute advantages $A_t = G_t - V_\phi(s_t)$\;
    Update critic $V_\phi$ by minimizing 
    $L_{\text{critic}}(\phi)$
    Eq.~\ref{eq:critic_loss}.\;
    Update policy $\pi_\theta$ by minimizing $L_{\text{policy}}(\theta)$
    Eq.~\ref{eq:policy_loss}.\; 
    % \tcp*{where $\beta>0$ is the entropy coefficient}
    Log metrics and save model if needed\;
    \If{early stopping criteria satisfied}{
        \textbf{break}\;
    }
}
\end{algorithm}

\paragraph{Output:}
The model outputs a vector of real-valued scores (logits or log-probabilities) over the action space: 
\begin{itemize}
    \item For the FC model: $\text{Output}_{\mathrm{FC}} \in \mathbb{R}^{144}$
    \item For the OC model: $\text{Output}_{\mathrm{OC}} \in \mathbb{R}^{4 \times n}$
\end{itemize}
During inference, these outputs are converted to a probability distribution over actions, from which the final action is selected either by sampling (stochastic policy) or by taking the action with the highest probability (greedy policy).

\subsection{Algorithm Implementation}

% We employ the Advantage Actor-Critic (A2C) algorithm by~\cite{mnih2016asynchronous}, using a Transformer-based architecture for both policy and critic networks. The preprocessed state representations described above are used as inputs. The policy network produces a probability distribution over actions (from the output logits), while the critic estimates the state-value function $V(s)$. Training proceeds by optimizing both networks jointly, enabling the agent to learn both the underlying rule structure and an optimal action-selection policy.

{We employ the Advantage Actor-Critic (A2C) algorithm (by~\cite{mnih2016asynchronous}) using a Transformer-based architecture for both the policy and critic networks. At each step, the environment $\mathcal{E}$ provides the agent with the current state $s_t$, a mask $m_t$ indicating invalid actions, and the valid action set ${v}_t$. The policy $\pi_\theta$ outputs a categorical distribution over ${v}_t$, from which the agent samples an action $a_t$. Executing $a_t$ in $\mathcal{E}$ yields the next state $s_{t+1}$ and reward $r_t$.}

Training proceeds by computing the discounted return
\begin{equation}
  G_t = \sum_{k=0}^{T-t-1} \gamma^{k}\, r_{t+k},
  \label{eq:returns}
\end{equation}
and the corresponding advantage
\begin{equation}
  A_t = G_t - V_\phi(s_t),
  \label{eq:advantages}
\end{equation}
which is normalized within a batch for variance reduction. The critic parameters $\phi$ are updated by minimizing the mean squared error
\begin{equation}
  L_{\text{critic}}(\phi) = \tfrac{1}{N}\sum_{t} \big(G_t - V_\phi(s_t)\big)^2,
  \label{eq:critic_loss}
\end{equation}
while the policy parameters $\theta$ are optimized with an entropy-regularized objective
\begin{equation}
  L_{\text{policy}}(\theta) = -\tfrac{1}{N}\sum_{t}
  \big(\log \pi_\theta(a_t \mid s_t)\,A_t + \beta\,H_t\big),
  \label{eq:policy_loss}
\end{equation}
where $H_t$ is the categorical entropy of $\pi_\theta(\cdot \mid s_t, m_t, {v}_t)$ and $\beta>0$ controls exploration.}
% \christo{included equations and updated the explanation.}

\subsection{Training and Evaluation Protocol}

The models are trained for a maximum of 10,000 episodes, with early stopping triggered once all three evaluation metrics are achieved.

\begin{table}[ht]
\centering
\begin{tabular}{l|l}
\hline
\textbf{Parameter} &  \textbf{Value} \\
\hline
$\alpha$ (Learning rate ) & $1 \times 10^{-5}$ \\
$\gamma$ (Discount factor) & $0.001$ \\
$\epsilon_{start}$  & $0.99$ \\
$\epsilon_{end}$ & $0.0001$ \\
$\epsilon_{decay}$ & $200$ \\
Batch size & 1 \\
n (Initial objects) & 9 \\
\hline
\end{tabular}
\label{tab:hyperparameters}
\caption{Training hyperparameters used in the reinforcement learning experiments.
{$\alpha$ controls the step size in gradient descent updates;
$\gamma$ determines the importance of future rewards;
$\epsilon_{\text{start}}$, $\epsilon_{\text{end}}$, and $\epsilon_{\text{decay}}$
define the $\epsilon$-greedy exploration schedule;
Batch size indicates the number of episodes per parameter update;
$n$ represents the number of initial objects placed on the game board.}}
\end{table}

\subsection{Evaluation Metrics}

% \pk{I think it is important that we use lower case for the individual values of the x* metrics, and we use upper case for the median of the set of observed values. SO there are in principle the data compared in the Spearman work, which are upper case numbers, for each of the many cases. And when you compare individual rules, you are using the set of lower case values. Christo -- please check this, and ask me to review it when you are  done.  }

\begin{enumerate}
    \item \textbf{e-star-mean {($e^\star_{mean}$)}:} 
    % \hao{also mention the variable notation in Row 1 of Table 3 here? same for other items}\christo{done, included mathematical notations here and as captions near table}
    The first episode, after which, for a window of size $W_{\mathrm{mean}}$,  the average error rate does not exceed the threshold $T_{\mathrm{mean}}$.
    \[
    e^\star_{\text{mean}} 
    = \min \Bigg\{\, t \;\Bigg|\; 
    \frac{1}{W_{\text{mean}}} \sum_{k=0}^{W_{\text{mean}}-1} E_{t+k} \;\leq\; T_{\text{mean}} \,\Bigg\},
    \]
    where $E_t$ denotes the error rate at episode $t$.
    \item \textbf{e-star-max {($e^\star_{max}$)}:} The first episode, after which, for 
    %within 
        a window of size $W_{\mathrm{max}}$, 
        %where 
        the maximum error rate does not exceed $T_{\mathrm{max}}$.

    \[
    e^\star_{\text{max}} 
    = \min \Bigg\{\, t \;\Bigg|\; 
    \max_{0 \leq k < W_{\text{max}}} E_{t+k} \;\leq\; T_{\text{max}} \,\Bigg\},
    \]
    \item \textbf{m-star {($m^\star$)}:} The first {step}, after which,  for a window of size {$W_{\mathrm{m\_star}}$},  all moves are successful.
    % \blue{
    % \[
    % m^\star 
    % = \min \Bigg\{\, t \;\Bigg|\; 
    % M_{t+k} = 1 \quad \forall \, 0 \leq k < W_{m^\star} \,\Bigg\}.
    % \]
    %  where $M_{t}=1$ if all moves in episode $t$ are successful and $M_{t}=0$ otherwise
    % }
    \[
        m^{\star} \;=\;
        \min \Bigl\{\, m \in \mathbb{N} \;:\;
           \forall i \in [\,m,\, m + W_{m^{\star}} - 1\,],\;
           \text{response\_code}(i) = \textit{'A'}
        \Bigr\}.
    \]
    where $\text{response\_code}(i)$ denotes the status code associated with move $i$ (see Sec.~\ref{sec:response-codes}).
    
\end{enumerate}

%\rc{i made edits to the definitions. I think these are clearer. Are they correct?}
%\blue{christo: they are more clear}
% \orange{Christo: I have added mathematical notations for the metrics. Are they correct and is it better to keep them? }
% \pk{I think they are correct. I would not have used $M$ as the name for the flag here, but it is technically ``not incorrect". 9/4/2025 10:42:12 AM}
% \christo{I have updated the eqn for m-star \pk{thanks 9/6/2025 1:49:16 PM}} 

\begin{figure}[htb]
    \centering
    \includegraphics[width=\columnwidth]{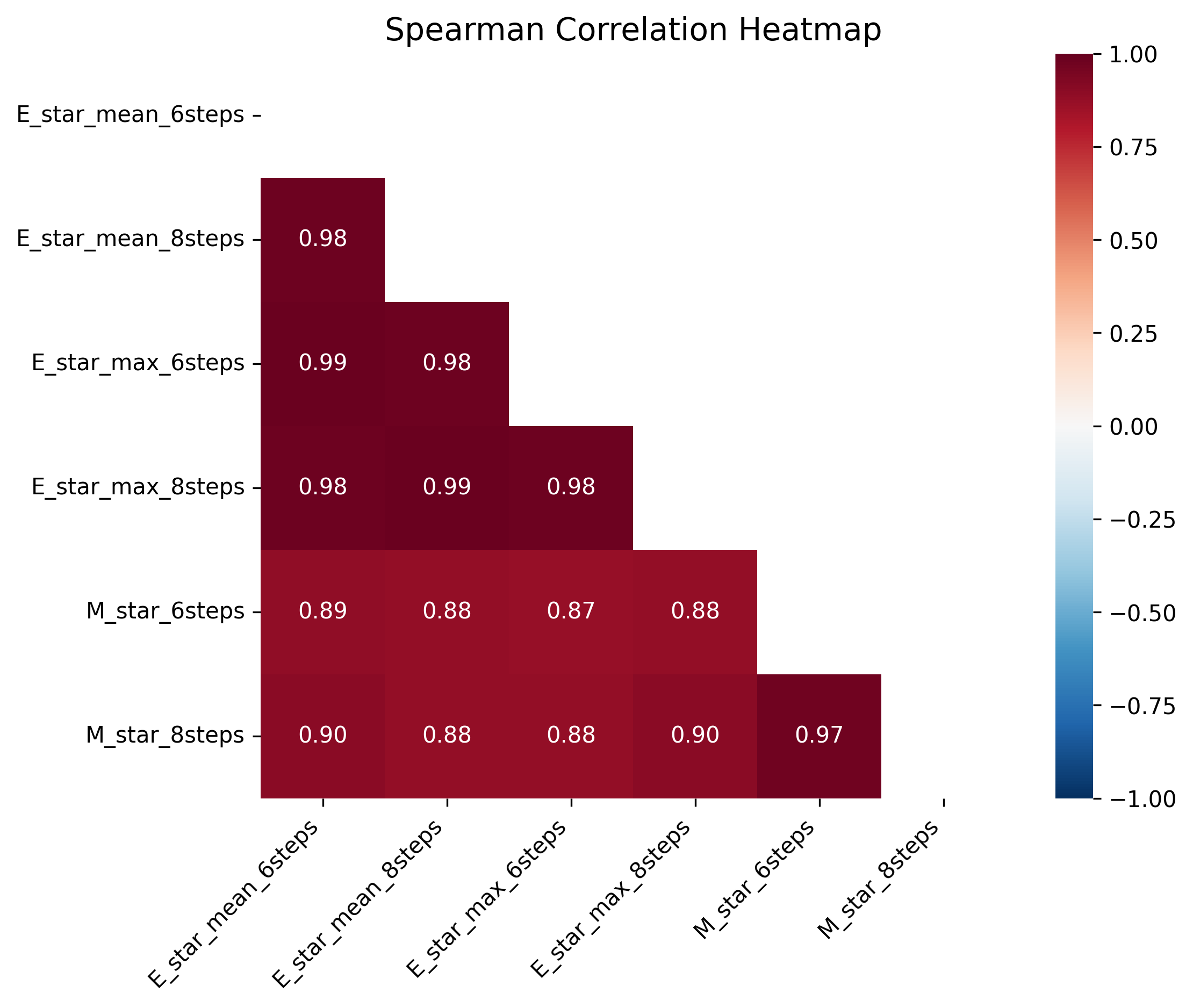}
    \caption{Heatmap of Spearman correlations between evaluation metrics.}
    \label{fig:spearman_heatmap}
\end{figure}

We find that the evaluation metrics are substantially  consistent but by no means identical.  The episode window-based methods agree extremely  well on the relative difficulty of rules, the $m^\star$ methods also show strong internal agreement. {We denote per-run metrics as $e^\star_{\text{mean}}$, $e^\star_{\text{max}}$, and $m^\star$, while their aggregated counterparts—computed as the median across runs—are denoted by $E^\star_{\text{mean}}$, $E^\star_{\text{max}}$, and $M^\star$.} .Together the two classes of metrics exhibit strong correlation across experiments on 18 different rules. Fig.~\ref{fig:spearman_heatmap} presents the heatmap of Spearman correlations, illustrating the relationships between the three metrics when evaluated across 18 rules using both 6-step and 8-step memory windows. The experimental data supporting this presentation are given in Table~\ref{tab:spearman-table}. 
% \hao{(WHEN?) please remember to add "." after \ ref{XXX} or \ cite{XX}, please check all paragraphs. Also m* refers to m-star above but it is never clarified explicitly. In Table 3 it shows M* not m*?}

% \rc{I used input  for the file with the table. It should look all right when we go back to one column. }

% \begin{table}[H]
% \centering
% \caption{\label{tab:spearman-table}Spearman Correlation Matrix}
% \centering
% \begin{tabular}[t]{l|cccccc}
% \hline
% Variable & $e^\star_{mean}6$ & $e^\star_{mean}8$ & $e^\star_{max}6$ & $e^\star_{max}8$ & $M^\star_{6}$ & $M^\star_8$ \\
% \hline
% e^\_star\_mean\_6steps & 1.00 & 0.98 & 0.99 & 0.98 & 0.89 & 0.90\\
% \hline
% e^\_star\_mean\_8steps & 0.98 & 1.00 & 0.98 & 0.99 & 0.88 & 0.88\\
% \hline
% e^\_star\_max\_6steps & 0.99 & 0.98 & 1.00 & 0.98 & 0.87 & 0.88\\
% \hline
% e^\_star\_max\_8steps & 0.98 & 0.99 & 0.98 & 1.00 & 0.88 & 0.90\\
% \hline
% M^\_star\_6steps & 0.89 & 0.88 & 0.87 & 0.88 & 1.00 & 0.97\\
% \hline
% M^\_star\_8steps & 0.90 & 0.88 & 0.88 & 0.90 & 0.97 & 1.00\\
% \hline
% \end{tabular}
% \end{table}

\begin{table}[H]
\centering
\caption{Spearman Correlation Matrix.
Here, {$E^\star_{\text{mean},6steps}$} refers to the $E^\star_{\text{mean}}$} for experiments with 6-step memory windows. 
% \christo{can we make the column names equal to the row names , since they are the same. ? I have updated the row and column names and made it to E* instead of e* since we are using medians. } \pk{looks OK to me 9/6/2025 1:48:35 PM} }
\label{tab:spearman-table}
\centering
\begin{tabular}[t]{l|cccccc}
\hline
Variable & $E^\star_{\text{mean},6steps}$ & $E^\star_{\text{mean},8steps}$ & $E^\star_{\text{max},6steps}$ & 
$E^\star_{\text{max},8steps}$ & $M^\star_{6steps}$ & $M^\star_{8steps}$ \\
\hline
$E^\star_{\text{mean},6steps}$ & 1.00 & 0.98 & 0.99 & 0.98 & 0.89 & 0.90\\
\hline
$E^\star_{\text{mean},8steps}$ & 0.98 & 1.00 & 0.98 & 0.99 & 0.88 & 0.88\\
\hline
$E^\star_{\text{max},6steps}$ & 0.99 & 0.98 & 1.00 & 0.98 & 0.87 & 0.88\\
\hline
$E^\star_{\text{max},8steps}$ & 0.98 & 0.99 & 0.98 & 1.00 & 0.88 & 0.90\\
\hline
$M^\star_{6steps}$ & 0.89 & 0.88 & 0.87 & 0.88 & 1.00 & 0.97\\
\hline
$M^\star_{8steps}$ & 0.90 & 0.88 & 0.88 & 0.90 & 0.97 & 1.00\\
\hline
\end{tabular}
\end{table}

% \begin{table}[H]
%     \centering
%     \caption{Spearman correlation matrix of evaluation metrics.}
%       \label{tab:spearman_correlation}
%     \resizebox{\textwidth}{!}{%
%     \begin{tabular}{lrrrrrr}
%         \toprule
%         Metric & e\_star\_mean\_6steps & e\_star\_mean\_8steps & e\_star\_max\_6steps & e\_star\_max\_8steps & M\_star\_6steps & M\_star\_8steps \\
%         \midrule
%         e\_star\_mean\_6steps & 1.00 & 0.98 & 0.99 & 0.98 & 0.89 & 0.90 \\
%         e\_star\_mean\_8steps & 0.98 & 1.00 & 0.98 & 0.99 & 0.88 & 0.88 \\
%         e\_star\_max\_6steps  & 0.99 & 0.98 & 1.00 & 0.98 & 0.87 & 0.88 \\
%         e\_star\_max\_8steps  & 0.98 & 0.99 & 0.98 & 1.00 & 0.88 & 0.90 \\
%         M\_star\_6steps       & 0.89 & 0.88 & 0.87 & 0.88 & 1.00 & 0.97 \\
%         M\_star\_8steps       & 0.90 & 0.88 & 0.88 & 0.90 & 0.97 & 1.00 \\
%         \bottomrule
%     \end{tabular}
%     }
% \end{table}

\section{Experiments}

The experiments using the GOHR environment encompass two primary configurations:

\begin{enumerate}
    \item \textbf{Rule-based configuration:} In this configuration, a rule file specifying the rule definition is provided to the game server.
    \item \textbf{Trial-list-based configuration:} In this configuration, a trial list is supplied, specifying both the rule set and the initial board generation process. This method also supports transfer experiments. 
\end{enumerate}

\subsection{Independent Rule Experiments }

Eighteen different rule files were considered for these experiments . The rules used are listed in Table~\ref{tab:rules-based exp} . Each rule was used to train a model independently, in order to assess model performance on isolated rule learning.

% \subsection{Analysis of Rule Difficulty}

For the analysis of rule difficulty, we initially attempted to analyze each rule separately. However, slight variations in metrics and the substantial overlap between metrics made it difficult to rank the rules clearly. Therefore, we decided to decompose the rules into their individual 
%aspects, 
{properties} (see Table~\ref{tab:rule-aspect-mapping}), which enabled us to better understand how the model performs for each {property} and identify their respective strengths and weaknesses.

We have identified the following {properties} for analysis:
\begin{enumerate}
\item \textbf{Quadrant\_to\_bucket\_mapping}: Mapping of pieces in a specified quadrant to a particular bucket.
\item \textbf{Proximity}: Removing pieces based on proximity to any bucket. ``Farthest'' indicates pieces at the center of the board, while ``nearest'' indicates pieces in any of the corners.
\item \textbf{Reading\_order}: Removing pieces in reading order (or reverse reading order), that is, from left to right and top to bottom (or right to left and bottom to top).
\item \textbf{Feature\_to\_bucket\_mapping}: Mapping each color/shape to any bucket.
\item \textbf{Feature\_ordering}: Removing one piece of each color/shape in order (e.g., blue after red, black after blue).
\item \textbf{All\_pieces\_of\_feature}: Removing all pieces of one color/shape, then proceeding to the next.
\item \textbf{Bucket\_ordering}: Assigning pieces to a specified order of buckets.
\item \textbf{Conditional}: Skipping pieces based on specified conditions.
\end{enumerate}

All rules can be characterized as either one of these properties or a combination of multiple properties.

\begin{table}[htb]
\centering
\caption{Mapping of Rules to Their Corresponding {Properties}}
\label{tab:rule-aspect-mapping}
\begin{tabularx}{\textwidth}{lX}
\toprule
\textbf{Rule Name} & \textbf{Properties} \\
\midrule
cm\_RBKY & Feature\_to\_bucket\_mapping \\
sm\_csqt & Feature\_to\_bucket\_mapping \\
allOfColOrd\_BRKY & All\_pieces\_of\_feature \\
allOfShaOrd\_qcts & All\_pieces\_of\_feature \\
col1Ord\_BRKY & Feature\_ordering \\
col1Ord\_KRBY & Feature\_ordering \\
colOrdL1\_BRKY & Feature\_ordering + Conditional \\
sha1Ord\_qcts & Feature\_ordering \\
shaOrdL1\_qcts & Feature\_ordering + Conditional \\
col1OrdBuck\_BRKY0213 & Feature\_to\_bucket\_mapping + Feature\_ordering \\
sha1OrdBuck\_qcts0213 & Feature\_to\_bucket\_mapping + Feature\_ordering \\
ordL1 & Reading\_order \\
ordRevOfL1 & Reading\_order \\
ordL1\_Nearby & Reading\_order + Proximity \\
ordRevOfL1\_Remotest & Reading\_order + Proximity \\
cw & Bucket\_ordering \\
ccw & Bucket\_ordering \\
cw\_0123 & Bucket\_ordering \\
cw\_qn2 & Quadrant\_to\_bucket\_mapping + Bucket\_ordering \\
quadNearby & Quadrant\_to\_bucket\_mapping \\
quadMixed1 & Quadrant\_to\_bucket\_mapping \\
\bottomrule
\end{tabularx}
\end{table}

\subsubsection{Difficulty Analysis of Rule {Properties}}

We observed a clear trend in the difficulty of learning each {rule properties}. The difficulty {is largely determined by the degree of abstraction in the feature underlying each property. } 
%primarily depends on the level of abstraction of the feature used in the aspects.
Even when considering features at the same level of abstraction, mapping to buckets is generally easier to learn than ordering based on features.

Based on our experiments and observations, {we establish the relative ordering of difficulty for rule properties in the FC and OC models, providing insight into how each representation influences learning performance.} The experiment results can be found in  Tables~\ref{tab: aspects_FC} and ~\ref{tab: aspects_OC}. When comparing {properties}, we primarily relied on $M^\star$ values. $m^\star$ is a step-level measure, defined directly in terms of whether individual moves obey the rule properties.
% \pk{Obey the aspect sounds very strange to me. Why not change it everywhere to ``obey the rule.''? I think you have used aspect in a sensible way to refer somehow to one or another of the concepts in the rule. But one has to obey ``all of the rule'' so it is hard to define what it means to not obey one aspect. I see that later on you  also use aspect to refer to the several ``difficulty classes.'' Perhaps better to just call them ``difficulty classes''} 
This makes it straightforward to compute, even for properties that are not well-defined on every move (for example, bucket\_ordering rules, where correctness/performance can only be evaluated after the first bucket placement has established a reference point, or conditional properties, which only apply when specific preconditions are met).

By contrast, $e^\star_{mean}$ and $e^\star_{max}$ are episode-level measures that require computing per-episode correctness rates. For %aspects 
% \pk{concepts }
properties that only become meaningful in certain moves, many episodes provide few opportunities to evaluate them, making these statistics noisier and harder to interpret. 
% \hao{what are the results that lead to the conclusions below. need to refer to some concrete tables, figures, etc. when discussing the results. Or, if you do not want to put results in this section, you can simply say that based on our preliminary results we find the following.}\christo{refered the table in the paragraph above.}

\paragraph{FC Model:}
In the FC model, positional properties are the easiest to learn (lowest level of abstraction), followed by properties that depend on piece features (higher level of abstraction).
\begin{enumerate}
    \item \textbf{Easiest properties}: Quadrant\_mapping, proximity, and reading\_order were the most accessible properties for the FC model to learn. This can be attributed to their primary dependence on piece positions, which the FC model captures most effectively. Among these, quadrant\_mapping was the easiest, followed by proximity with nearly identical learning curves, while reading\_order proved more challenging.

    \item \textbf{Moderate Difficulty}: Feature\_to\_bucket mapping follows next, depending on individual piece features such as color or shape being mapped to output buckets. The all\_pieces\_of\_feature property, which also depends on piece features, exhibits similar difficulty.

    \item \textbf{Higher Difficulty}: Bucket\_ordering proved more challenging, as it depends on an abstract ordering of buckets that is difficult for the FC model representation to capture.

    \item \textbf{Most Challenging}: Feature\_ordering and conditional properties represent the most challenging categories for the FC model, exhibiting very slow learning curves.

\end{enumerate}

\paragraph{OC Model:}

While the FC model showed significant differences in difficulty between properties, the OC model exhibited relatively smaller differences in difficulty and learning curves between properties.

\begin{enumerate}
    \item \textbf{Easiest properties}: Feature\_to\_bucket mapping, all\_pieces\_of\_feature, and quadrant\_mapping were most accessible for the model to learn. These properties depend on the lowest level of feature abstraction, directly utilizing color, shape, or x,y coordinates. These properties demonstrated nearly identical learning curves with minimal differences. Notably, quadrant\_mapping was the most challenging among these three, which can be attributed to its dependence on two features (x and y coordinates) rather than a single feature.

    \item \textbf{Moderate Difficulty}: Bucket\_ordering proved slightly more challenging than the easiest properties.

    \item \textbf{Higher Difficulty}: Reading\_order and proximity exhibited slower learning curves compared to the above properties. This can be attributed to the higher level of feature abstraction, as the model must derive positional information from the provided x and y coordinates.

    \item\textbf{Most Challenging}: Consistent with the FC model, feature\_ordering and conditional properties remained the most challenging for the OC model to learn, exhibiting very slow learning curves.

\end{enumerate}

\subsubsection{Rule Difficulty Analysis}

Having established the relative difficulty ordering for individual rule properties, we now examine the difficulty ordering of complete rules. 

{Rules ordered by increasing difficulty are presented in Tables~\ref{tab:FC_difficulty} and~\ref{tab:OC_difficulty}. The rules are grouped into categories according to the ranges of their metric values. While the ordering of rules within each category is somewhat ambiguous and may vary depending on the chosen metric, we consistently observe a clear relative ordering across categories. This relative structure remains stable across all three metrics.}

% We observed that when two aspects are combined, both of their learning curves become slower.\blue{In the table \ref{tab: aspects_FC} and table \ref{tab: aspects_OC} we can observe that when combined with proximity, the performance of reading order decreases. }
% \hao{what are the results that lead to the conclusions below. need to refer to some concrete tables, figures, etc. when discussing the results. }
% \christo{has included references to table in the paragraph above}

\paragraph{FC Model}

\begin{enumerate}
    \item \textbf{Highly Learnable}: The most accessible rules are the quadrant\_mapping rules: quadNearby and quadMixed1.

    \item \textbf{Moderately Learnable}: Rules based on proximity and reading\_order follow: ordL1\_Nearby, ordL1, ordRevOfL1, and ordRevOfL1\_Remotest.

    \item \textbf{Challenging}: Rules depending on feature\_to\_bucket mapping and all\_pieces\_of\_feature properties include cm\_RBKY, sm\_csqt, allOfShaOrd\_qcts, and allOfColOrd\_BRKY.

    \item \textbf{More Challenging}: The rules col1OrdBuck\_BRKY0213 and sha1OrdBuck\_qcts0213, which are primarily feature\_to\_bucket mapping rules, show slower learning due to the additional feature ordering component. The ccw and cw rules demonstrate comparable difficulty to these combined rules.

    \item \textbf{Highly Challenging}: Rules depending on feature\_ordering properties (sha1Ord\_qcts and col1Ord\_BRKY) are very difficult to learn.

    \item \textbf{Most Challenging}: When conditional properties are added to feature\_ordering, the rules become most challenging for the model. This category includes colOrdL1\_BRKY and shaOrdL1\_qcts.

\end{enumerate}

\begin{figure}[htbp]
  \centering
  \includegraphics[width=\linewidth]{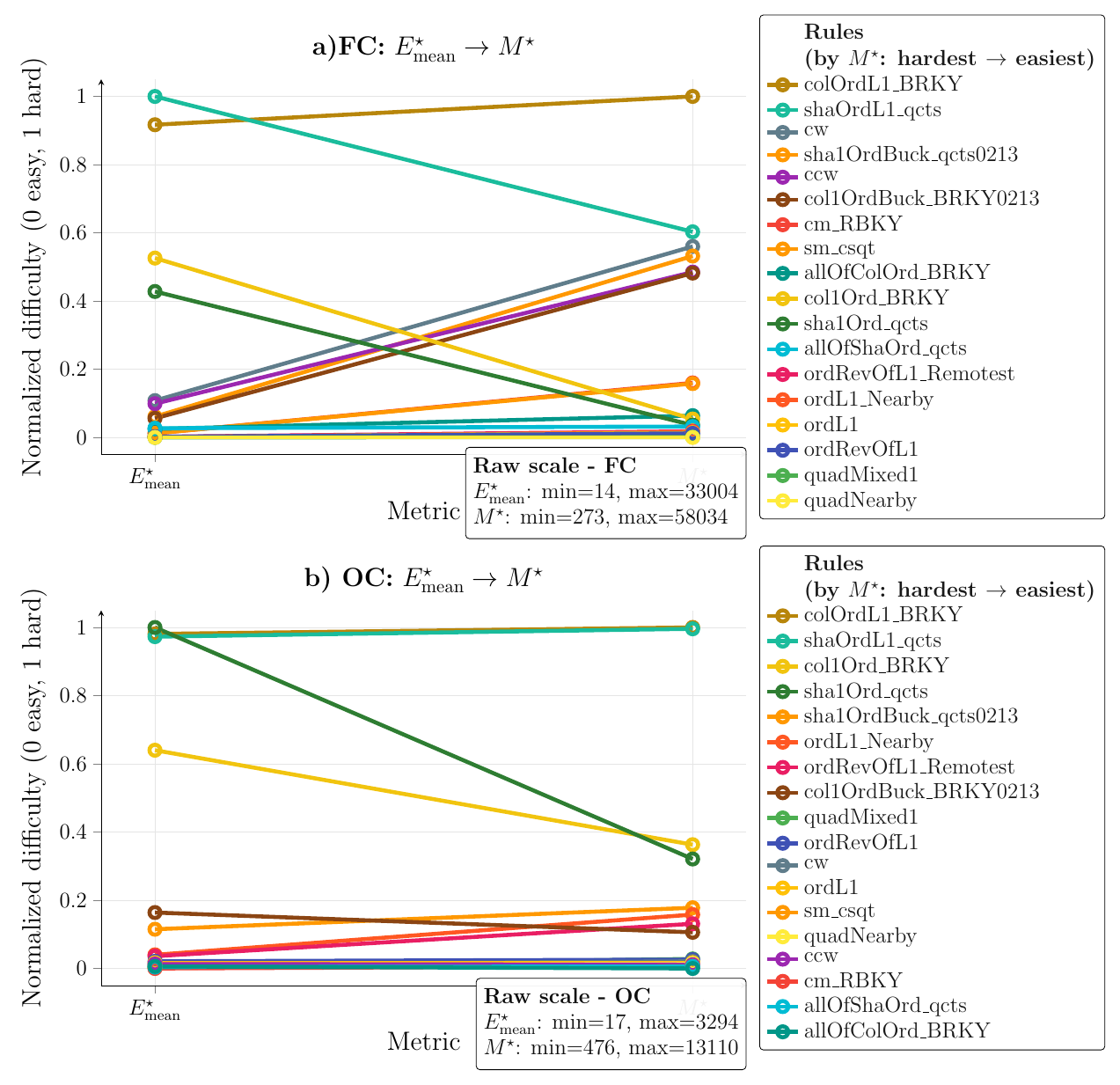} % path to compiled PDF
  \caption{Crossings plots for rule difficulty between different metrics. 
    Each line connects a rule's normalized difficulty score in the  $E^\star_{mean}$ (left) to the $M^\star$ (right). 
    Labels are displayed in the order of their $M^\star$ values ;color of rule label corresponds to the color of the line it represents; raw scales (min, median, max per model) are shown inside each panel. A scale including the minimum, median and max values of each metrics is included. 
    (a) shows the comparison of metrics in FC model and (b) shows the comparison of the metrics in OC model.} 
    % \hao{Maybe use larger fonts for ALL text in the figure?}\blue{christo: updated the figure, there was a mistake. Changed to comparison of metrics. I think it is more intuitive
    % \hao{Thanks! By all text, I mean you also need to use larger font for the xlabel ylabel xtick ytick of the plot.}
    % \christo{Updated the figure and enlarged the fonts. 9/6/25 12:15}}
  \label{fig:fc_oc_diff_analysis}
\end{figure}

\paragraph{OC Model}

In the OC model, rule difficulties are more closely clustered, making ordering more difficult based on final metrics, with apparent crossings between metrics.
\begin{enumerate}
    \item \textbf{Learnable}: The most accessible rules include allOfColOrd\_BRKY, allOfShaOrd\_qcts, cm\_RBKY, sm\_csqt, quadNearby, quadMixed1, cw, ccw, ordL1 and ordRevOfL1. These rules primarily rely on properties such as feature\_to\_bucket mapping, all\_pieces\_of\_feature removal, quadrant\_mapping, and bucket\_ordering. Among them, ordL1 and ordRevOfL1, which are based on reading\_order, show slightly higher difficulty but remain close to the others in this group and can still be considered part of the learnable category.

    % \item \textbf{Learnable}: Rules ordL1 and ordRevOfL1, which depend on reading\_order, exhibit difficulty levels very close to the above rules and can be considered part of the learnable category.

    \item \textbf{Moderately Challenging}: When proximity is added to the reading properties, rules become more challenging to learn, including ordRevOfL1\_Remotest and ordL1\_Nearby. The rules col1OrdBuck\_BRKY0213 and sha1OrdBuck\_qcts0213, which combine feature\_to\_bucket mapping with feature\_ordering, demonstrate comparable difficulty to reading+proximity rules.

    \item \textbf{Most Challenging}: Similar to the FC model, rules depending on feature\_ordering are the most difficult, including col1Ord\_BRKY, sha1Ord\_qcts, colOrdL1\_BRKY, and shaOrdL1\_qcts. The addition of conditional properties makes colOrdL1\_BRKY and shaOrdL1\_qcts the most challenging rules for the model to learn.

\end{enumerate}
\subsubsection{Key Observations}

{
In Fig.~\ref{fig:fc_oc_diff_analysis}(a) we observe numerous crossings between the %$e^\star_{\text{mean}}$
$E^\star_{\text{mean}}$ and $M^\star$ rankings. Most crossings occur within or between closely related rule properties categories (as defined in the Rules/Rule Property Difficulty Analysis), indicating that in the FC representation the relative difficulty is metric-sensitive: rules can swap order depending on whether we emphasize learning speed ($e^\star_{\text{mean}}$) or final convergence ($m^\star$). This effect is especially visible among feature-dependent 
%categories
properties (feature\_to\_bucket, all\_pieces\_of\_feature, feature\_ordering), whereas the easiest positional rule properties (e.g., quadrant/proximity/reading\_order) and the hardest feature\_ordering/conditional rule properties tend to preserve their extremes of the ranking.} 
% \hao{Please use consistent style, either Figure XX or Fig. XX throughout the paper.} \christo{done, modified all figure references to Fig.}

{By contrast, Fig.~\ref{fig:fc_oc_diff_analysis}(b) shows substantially fewer crossings, yielding a more stable rule ordering across metrics. This aligns with our earlier observation that 
% OC exhibits smaller between-difficulty gaps: 
 OC representation exhibits smaller gaps between rule properties:feature\_to\_bucket, all\_pieces\_of\_feature, and quadrant\_to\_bucket\_mapping cluster together on the
%easy
easier side, while feature\_ordering and conditional rules remain hardest, with the same relative ordering under both metrics. In short, OC provides a more coherent notion of difficulty across measures, whereas FC shows larger metric-driven variability, primarily within or at the boundaries of related categories.}

The complete results of the independent rule experiments are available at: \url{https://tinyurl.com/4vfjp6yn}.

% \christo{description mentioning figure. I have included learning speed ($e^\star_{\text{mean}}$) or final convergence ($M^\star$). Please let me know if this is correct approach.}\pk{Yes, I think this is a good selection. I changed ``aspect' to ``difficulty'' here 9/4/2025 10:41:07 AM}
% \christo{I have changed all the occurence of ``aspect`` to ``property``. Because difficulty classes may be confusing. Please let me know if further change is required 9/5/2025 15:35 AM. \pk{I think this would be too small in print, but people are almost surely reading on a computer, so the small detailes are OK. 9/6/2025 1:41:06 PM}}

% \blue{\textbf{Effect while merging aspects}:}We observe distinct behaviors when combining different types of aspects. When merging proximity and reading order aspects (two independent aspects), the difficulty increases significantly. However, when merging feature\_ordering and feature\_to\_bucket mapping, the resulting difficulty falls in between the difficulty of those individual aspects; this occurs because these two aspects overlap and exhibit a compound effect that differs from the pattern observed when merging independent aspects.

\subsection{Transfer Effect Experiments}

The transfer effect of rules were studied using trial-list based configuration. 
Each trial-list file contains four trial-list combinations, with each row representing a specific configuration. These experiments are designed to observe the \emph{transfer effect}—that is, whether learning is facilitated when the model is exposed to rule combinations after training on individual components. 

% \rc{why not change to A+B for purposes of the paper? }
% \pk{Thanks for changing to +. I took away the blue 9/4/2025 10:20:22 AM}
Consider four independent rules, A, B, C, and D. A+B represents a rule that is a combination of A and B. Two types of transfer experiments were performed:

\begin{enumerate}
    \item \textbf{Sequential Transfer:} In this sequence, the agent is first trained on two rules separately (A and B), then on their combination (A+B), followed by exposure to a combination of two entirely different rules (C+D). This protocol is intended to test whether prior training on the component rules facilitates faster or more robust learning when faced with their combination, and to evaluate the extent of positive or negative transfer when learning new combinations thereafter.

    \item \textbf{Partial Transfer:} Here, the agent is trained on two rules in sequence (e.g., A and C), followed by training on a combination that includes one previously learned rule and one novel rule (A+B and C+D). This design allows us to analyze whether experience with one rule aids learning its combination with an unfamiliar rule, as well as to compare transfer effects relative to the sequential transfer scheme.

\end{enumerate}

For both protocols, $\epsilon$ is re-initialized to $\epsilon_{\text{start}}$ at the start of each phase to promote exploration and prevent premature convergence. Rule combinations used in these protocols are listed in Tables~\ref{tab:sequential_transfer} and ~\ref{tab:partial_transfer}. Rule descriptions are available in Table~\ref{tab:rules-desc}. 
% \hao{need to fix the broken Table ref.}\christo{fixed}

Each experiment was conducted over five independent runs with different random seeds. For each run, the evaluation metrics $e^\star_{\text{mean}}$, $e^\star_{\text{max}}$, and $m^\star$ were recorded. The final results were obtained by aggregating across runs using the median, reported as \emph{$E^\star_{\text{mean}}$}, \emph{$E^\star_{\text{max}}$}, and \emph{$M^\star$}, respectively.

{\subsubsection{Key Observations}}
From the experiments, we observe the following transfer effects (see Figs.~\ref{fig:transfer_all} for details):
% ~\ref{fig:fc_cm_cw},~\ref{fig:fc_cm_ordL1},~\ref{fig:fc_cm_cwqn2},~\ref{fig:fc_ordL1Nearby},~\ref{fig:oc_cm_cw},~\ref{fig:oc_cm_ordL1},~\ref{fig:oc_cwqn2},~\ref{fig:oc_ordL1Nearby}, and related plots for details):  

\begin{enumerate}
    \item \textbf{Full-component pretraining accelerates learning.} Compound rules (A+B) learn substantially faster when trained immediately after their component rules. In particular, when A+B follows pretraining on both A $\rightarrow$ B (A followed by B), the model converges quickly to the rule.  

    \item \textbf{Partial pretraining yields slower convergence and depends on order.} If A+B is trained after only one component (A or B) combined with an unrelated rule C, learning is slower than with full-component pretraining. Moreover, the convergence speed depends strongly on the specific rules and their order. For example, in the OC model with compound rule cw\_qn2 (cw + quadNearby), pretraining with cw $\rightarrow$ cm\_RBKY or quadNearby $\rightarrow$ ordL1 gives moderate improvements, but ordL1 $\rightarrow$ quadNearby leads to much faster convergence, nearly matching cw $\rightarrow$ quadNearby.  

    \item \textbf{Adding extra rules after both components harms transfer.} When A+B is trained after a sequence A $\rightarrow$ B $\rightarrow$ C, convergence is substantially degraded, resembling the effect of partial pretraining rather than full.  

    \item \textbf{Unrelated pretraining provides the weakest transfer.} If A+B is trained after unrelated rules C $\rightarrow$ D, convergence is the slowest, showing minimal benefit from prior training.  
\end{enumerate}

% \hao{so far we have e-star, E-star, e*, etc. probably better to be consistent} \pk{Lower case refers to individual values. Upper case is the median of the observed values }

The complete results of the transfer effect experiments are available at: \url{https://tinyurl.com/2dvv6f97}.
% The complete results of the independent rule experiments and transfer analyses are available at: \url{https://tinyurl.com/yd97zd4s}
% The complete set of results is available at:
% \url{https://tinyurl.com/hf47t9zc}. 
% \hao{This is too like and a bit overflow, maybe use tinyurl here?}

% \hao{we may want to summarize some findings here. otherwise this does not look like a CS tech report.}\christo{added the subsubsection "4.2.1 key observations"}

\subsection{Generalization Analysis}

We conducted experiments to evaluate the generalization ability of our models using the ‘train’ and ‘test’ modes available in the GOHR environment. In ‘train’ mode, pieces are restricted to a specified subset of positions, whereas in ‘test’ mode, at least one piece is placed in a position not used during training.

For our experiments, we adopted the checkerboard pattern (see Fig.~\ref{fig:checkerboard}). In ‘train’ mode, pieces were placed only on the white squares of the checkerboard, which served as the training set for the RL agent. In contrast, ‘test’ mode allowed placement across the entire board, thereby testing the model’s ability to generalize beyond the restricted training positions.

\begin{figure}[htb]
    \centering
    % \documentclass[tikz,border=6pt]{standalone}
% \usepackage{tikz}

% \begin{document}
\begin{tikzpicture}[x=0.7cm,y=0.7cm]
  % ----- Black cells by index (1..36; 1 is bottom-left, 6 bottom-right) -----
  \foreach \n in {2,4,6,11,9,7,14,16,18,23,21,19,26,28,30,35,33,31}{
    \pgfmathtruncatemacro{\c}{mod(\n-1,6)}       % 0-based column
    \pgfmathtruncatemacro{\r}{floor((\n-1)/6)}   % 0-based row from bottom
    \fill[black] (\c,\r) rectangle ++(1,1);
  }

  % ----- Grid & border (white squares are just the page background) -----
  \draw (0,0) grid[step=1] (6,6);
  \draw[line width=0.8pt] (0,0) rectangle (6,6);

  \foreach \r in {0,...,5}{
    \foreach \c in {0,...,5}{
      \pgfmathtruncatemacro{\idx}{\r*6+\c+1}
      \node[font=\scriptsize] at (\c+0.5,\r+0.5) {\idx};
    }
  }
\end{tikzpicture}
% \end{document} 
    \caption{Checkerboard pattern. The white squares represent positions used explicitly for training.}
    \label{fig:checkerboard}
\end{figure}

Our 
findings, to be reported elsewhere,
% results 
show clear differences between the FC and OC models. For FC models, the proportion of errors occurring in test mode (test-error ratio) generally ranged between 50–75\%. Notable exceptions were the rules cm\_RBKY and sm\_csqt, which reached ~90\%, and quadNearby and quadMixed1, which had near-complete failure with 99–100\% test errors. In contrast, the OC models showed more stable generalization, with most rules exhibiting test-error ratios around 49–55\%, although quadNearby and quadMixed1 again had the highest ratios (~65\%).

The increased error rates for quadNearby and quadMixed1 can be attributed to the nature of these rules, which depend heavily on the precise board locations of pieces. Since the models encode positions using raw $(x,y)$ values rather than quadrant indices, encountering unseen coordinates during testing substantially reduces performance.

Overall, we conclude that OC models exhibit stronger generalization by associating behavior with object-level features, whereas FC models are more limited, tending to memorize positional patterns rather than abstracting to object-level rules.

% \christo{added the section 'generalization analysis' which describes performance in the testing mode. \pk{ 9/6/2025 1:35:14 PM Christo - thanks very much for including this. It is the kind of observation that I have been hoping to see since we started looking at the OC representation. I think it can be another entire technical report. Added a few words in red.} }

\subsection{Similarity of shape and color rules}

During independent rule experiments we have assessed the performance of the model on isolated rules. An association that one would expect
% notice among the rules are how related or similar the 
is among the shape and color match rules. By definition, these rules differ only by using shapes instead of the colors, which are both encoded as one-hot encoding for the model; therefore, we expect those rules to be "similar". 

To verify if the shape and color rules were statistically similar, we have 
%conducted 
used the non-parametric Kruskal--Wallis Test. \\
% \begin{enumerate}
% \\
% \textbf{Kruskal-Wallis Test} \\
% \begin{itemize}
%     \item $H_0$: The distributions of the metric are the same for all groups. \\
%     \item $H_1$: At least one group differs in distribution (median) from the others.
% \end{itemize}

% \rc{I see you know how to make a box, so you should be able to do it for the second Thomson quote...:) }

\begin{tcolorbox}[colback=gray!10!white, colframe=black, title=Kruskal--Wallis Test]
\textbf{Null hypothesis ($H_0$):} The medians of the distributions of the metric are the same for all groups.\\[0.5em]
\textbf{Alternative hypothesis ($H_1$):} At least one group differs in (median) from the others.
\end{tcolorbox}

%     \item \textbf{Permutation Test} \\
%     $H_0$: The metric is exchangeable between groups; there is no difference in the distribution of the metric between groups. \\
%     $H_1$: The distribution of the metric differs between groups.

%     \item \textbf{Cliff's Delta} \\
%     Values range from $-1$ (all values in group 1 are less than group 2) to $1$ (all values in group 1 are greater), with $0$ indicating complete overlap.
% \end{enumerate}

Our analysis was conducted by comparing the similarity of the rules based on the three  metrics: $m^\star$, $e^\star_{mean}$, $e^\star_{max}$.
% m\_star\_values, e\_star\_mean\_vals and e\_star\_max\_vals .
 We conducted two types of analyses, and the details along with the associated statistical tests are presented below.
% The analysis files can be found at: \pk{the link does not show here. so the sentence is wrong. 9/4/2025 10:44:17 AM } \href{https://drive.google.com/drive/u/5/folders/1p2MYU8fON2sz6157oLR4AG5vQJPw-Rqr} The details of the analysis and associated tests are the results of our analysis is listed below:

\subsubsection{Comparison within the rule}

To assess whether each set (five runs) of experimental runs per rule for a rule produced similar outcomes, we applied Kruskal--Wallis test for all the three evaluation metrics. 
% \hao{give an example of what comparison you are doing.}

% all three statistical tests \rc{you only named one statistical test. What is the ``all three''?} across all three evaluation metrics. 

% For every rule, at least one metric produced a “True” result. %In particular, the Kruskal-Wallis test using m\_star\_values returned “True” for all rules, except for ColOrdL1\_BRKY with the FC model, where only e\_star\_max produced “True”. 

For every rule, at least one metric indicated that the runs were statistically indistinguishable; that is, the Kruskal--Wallis test failed to reject the null hypothesis for that metric. This shows that although there was variability across metrics, in each case there existed at least one evaluation dimension where the outcomes of the five runs could be considered ``not different.''

\subsubsection{Comparison between color and shape rules}

{We next compared the putatively identical color-based and shape-based rules, which differ only by using shape instead of color. {For example, we compared allOfColOrd\_BRKY vs. allOfShaOrd\_qcts, and col1Ord\_BRKY vs. sha1Ord\_qcts.}For each such pair, we ran the Kruskal--Wallis test on the three metrics. To improve stability, we aggregated seven sets of five runs (35 runs per rule), thereby reducing run-to-run variability and strengthening generalization.} 
% \hao{give an example of what comparison you are doing. what rule versus what rule?}\christo{examples highlighted in violet; 9/5/25 18:25}

Out of 30 comparisons (10 rule pairs $\times$ 3 metrics), 29 yielded $p$-values above 0.05, indicating no significant differences. The only exception was the pair \texttt{(col1Ord\_BRKY, sha1Ord\_qcts)} with the FC model under the $m^\star$ metric ($p = 0.036$). However, for this pair the other two metrics gave $p > 0.05$, so overall the rule pairs can be considered statistically similar (more precisely, ``not pairwise distinct'').  

\subsubsection{Effect of increasing the number of runs}

Because nonparametric tests such as 
Kruskal--Wallis 
% \hao{– or -? same for all mentions of Kruskal--Wallis in the paper.}\christo{used -- everywhere ; 9/5/25 18:25} 
are sensitive to sample size, we further investigated whether increasing the number of runs improves stability. We did this by pooling three sets (15 runs per rule) for the within-rule analysis. 

{
The proportion of ``not different'' outcomes increased from 66\% (40/60 comparisons) with 5 runs to 83\% (50/60 comparisons) with 15 runs. This demonstrates that a larger number of runs reduces variability and yields more stable conclusions. Importantly, the 5-run analyses still provide a valid baseline; the extended analyses serve as a robustness check that strengthens the results and shows that the expected similarities are more consistently recovered with larger sample sizes.  }

\subsubsection{Summary}

In summary: 
\begin{enumerate}
    \item The within-rule analysis using 5 runs revealed variability across metrics, but always included at least one metric where runs were statistically indistinguishable.
    \item The color-vs-shape comparison, using 35 runs per rule, confirmed that the expected pairs behave similarly, with 29 out of 30 comparisons showing no significant difference.
    \item  Increasing the number of runs in the within-rule analysis from 5 to 15 improved the proportion of ``not different'' outcomes from 66\% to 83\%, reinforcing the conclusion that limited run counts can increase variability, but the overall pattern of similarity holds.  
\end{enumerate}

The detailed results of the analysis can be found at: \url{https://tinyurl.com/55e4bxbz}.

% \url{https://tinyurl.com/2p63ydpb}

% \rc{is this usual -- to only refer to a web site? }
% \christo{christo: I am not sure, please let me know if we can include these table as csv in appendix} \pk{I think we can add supplementary materials as a bib item, and that can point to a web site  9/4/2025 10:45:51 AM}

\section{Multi-Dimensional Scaling}\label{Sec:MDS_Scaling}

The data on difficulty provides one dimension for understanding relations between rules. On the difficulty scale, rules that are ``near each other'' have something in common, which we have speculated about in the discussion above. But there are other ways of organizing the rules, based on the data we have collected. 

For example, each data point, characterized by a $(rule,algorithm)$ pair actually represents the median of five distinct observations, as the algorithm worked on the rule from a cold start five times, That set of five values contains information beyond its median. For example, the range is approximately a 94\% confidence interval for the median of the (unknown) distribution of that specific metric for all possible encounters of that rule with that algorithm 

A more rigorous way to explore the relation is to apply the Mann-Whitney test for equality of medians. When applied to two samples, it gives us the exact probability that the two observed sets of numbers could have come from the same distribution. The exact probability is calculated by computing the value of the specific Mann-Whitney U statistic, and asking how many of the values could be more extreme than the one that is observed.  We have calculated these {$p$-values}
% \hao{what's --values?}
% for 
% \violet{for the similarity between every possible pair of rule sets among the 18 total rule sets (See Table~\ref{tab:rules-based exp}). We computed the similarity between rule sets using the Mann-Whitney U test and recorded the p-values in a matrix.}
{for the similarity between every possible pair of rule sets among the 18 total rule sets (see Table~\ref{tab:rules-based exp}). The resulting $p$-values were recorded in a matrix.}
% \pk{Christo, can you fill in here about which cases you computed? I think when we have large data files we don't put in a link, but indicated that supplementary data can be found either on our github, or deposited at some data repository. PRof. Wang will be more familiar with the customs in CS. } \pk{The new purple looks OK to me. 9/4/2025 10:47:15 AM}
% \christo{ restructured the sentence to correct '--values' 9/5/2025 11:35}

\begin{figure}[t]
\centering
  \includegraphics[width=0.96\linewidth]{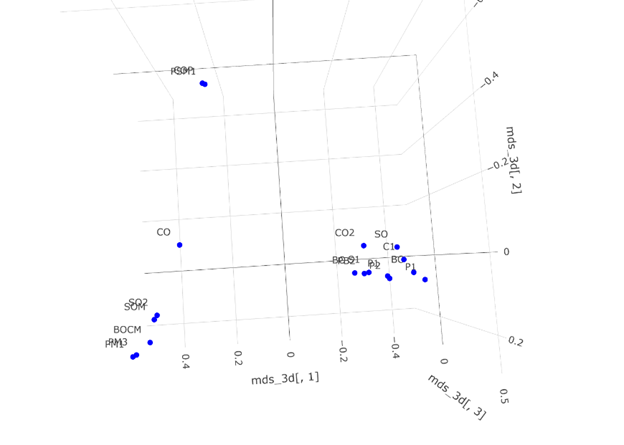} \hfill
  \caption {View of the first two principal dimensions of the MDS embedding of $p$-values as dissimilarity. Note the heavy cluster containing many different rules/}
  \label{fig:topview_of_MDS}
 \end{figure}

Specifically, we recognize that the $p$-value is a measure of similarity. Values close to $1$ are an indication that the sets of values could indeed come from the same distribution, while values close to zero are very unlikely, if both situations have the same distribution of difficulties.  For an initial look at the structure of the data, we compute a dissimilarity matrix $D_{i,j}=1-p_{MannWhitney}(i,j)$. We then used the R statistcal environment to transform this into a lower diagonal distance array. We then used the basic R tool for multidimensional scaling, to  look for  a $3-dimenional$ embedding of the data into a Euclidean space. The results are a rotatable plot, The first two principal dimensions of the analysis are shown in Fig.~\ref{fig:topview_of_MDS}. It is easy to see that there are a few clusters, but each of them  seems to contain ``too many'' different rules. 

\begin{figure}[ht!]
\centering
  \includegraphics[width=0.96\linewidth]{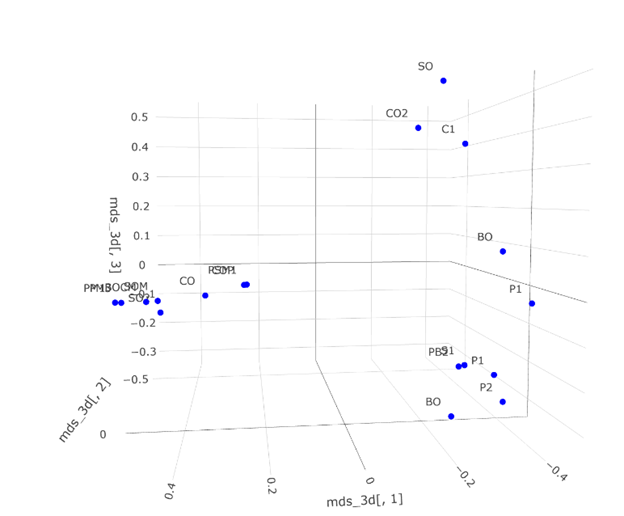} \hfill
  \caption {Perspective View of the first three  principal dimensions of the MDS embedding of $p$-values as dissimilarity. The figure has been rotated to resolve two clusters that each contain many point in the two dimensional view with the first two components. }
  \label{fig:frontview_of_MDS}
 \end{figure}

We continued the analysis, looking at a view using the third dimension as the vertical axis. In that view, shown in Fig.~\ref{fig:frontview_of_MDS} it is clear that the large cluster is resolved by differing values along the (now vertical) third axis. Future work will address the question of whether all of these clusters make sense both in terms of the specifics of the algorithm and in terms of the underlying concepts within the rules.

%$$$$$$

\

\

\section{Conclusion}

In this work, we investigated reinforcement learning in the Game of Hidden Rules (GOHR) environment—a setting where agents must infer and act based on hidden rule structures. We implemented and compared Feature-Centric and Object-Centric state representations within a Transformer-based Advantage Actor-Critic (A2C) framework, analyzing how different encoding %strategies 
choices
influence learning efficiency, transferability, and generalization. Our initial findings suggest that it may be possible to  define a space in which the various tasks ``discover such and such hidden rule A'' or ``building upon having discovered some rule A, discover some other rule B'' can be situated in such a way that proximity reflects their ``degrees of difficulty'' for ``artificial intelligence.'' While it was natural to expect that different forms of AI might situate these tasks differently in that space, we did not anticipate that different ways of measuring how difficult each  task T is, for system S, might yield different geometric or ordinal relationships. 

We close with a few conjectures about how this work might evolve with further study. 

% \hao{need better format to list these, e.g., with boldface like "\textbf{Conjecture 1.}" We can also split this section into several subsections?}\christo{made the main points like conjucture1, transferability bold 9/6/25 12:30 \pk{looks OK to me 9/6/2025 1:31:57 PM}}

\textbf{Conjecture 1.} Rules that are quite different in their concepts or conceptual complexity will not be ``near each other'' by reasonable metrics on the  space of rule difficulty. 
% \hao{this sentence is difficult to parse and sounds like circular definition (I am sure it is not meant to be). It basically says that "Rules that are quite different in difficulty will not be ``near each other'' in terms of difficulty. PAUL -- I changed it a little.}

\textbf{Transferability.} This operationalizes the notion of ``asking the subject to discover a second rule, after having succeeded in discovering a first rule.'' For a machine the process could involve some kind of freezing of all internal parameters when performance on the task corresponding to the first rule is judged adequate. In the present experiments, we have continued the learning process.  If having discovered the first rule makes discovering the second rule less difficult than it would be ab initio then, in some sense, the second rule is ``not very different'' from the first rule. We might  say that the first rule ``prepares the learner to discover the second rule.'' 

\textbf{Conjecture 2.} If each of two rules, A and B, are components of a 
%prepares the learner for a 
third rule C, then having learned both of them as concepts makes 
%learning 
C 
%even 
easier to discover. 
% \hao{what does "prepare" mean exactly here? maybe an example would help. Otherwise this sentence sounds like you are simply defining "prepare" as "having learned both of them makes learning C even easier to discover", which is not a conjecture.PAUL == I agree. I added some words and deleted others SEP 03 }

We note that the ``prepares'' relation may not be symmetric. In addition, there may be negative preparation relations, in which having learned rule A leaves the learner less able to discover rule C. 

\textbf{Complexity.} Rules may be complex in many ways. A rule may depend on only one feature – such as shape, or color. Or may depend on combinations of features. A rule may be stable over time, or depend on previous moves, etc.  This points to  a tempting third conjecture.

\textbf{Conjecture 3.} More complex rules always are more difficult to learn. 
% \hao{this conjecture is much clearer than the previous two}

The results reported here do not yet put a well-defined metric onto the space of rules.  Let us call that hypothetical space  a ``difficulty space.'' Conceivably, for many AI systems, if not all of them, the best achievable results may only be  a quasi-metric space, in which d(A,B) does not equal d(B,A). We anticipate that it may be possible to move towards establishing   a rigorous description of a machine’s intelligence, by measuring how far its performance  is from some large set of fiducial points in the space of difficulty. 

If  we are using rules that can be discovered both by machines and by humans, we may actually be able to situate a given human’s intelligence in that same space. Then – but we are almost in the realm of science fiction here – a digital twin machine learner 
%could 
might 
be used to select an optimal teaching schedule without the huge wasted effort of present-day educational experimentation. 

We stand at the edge of a transformative era in which machines, which have long supplemented our muscles, will supplement our brains in ways that we do not yet understand.  Laboratories in which computer scientists and psychologists work side by side, developing new instruments and metrics, have the potential to build the foundations of a new branch of science.  While we cannot predict that path of this new science, we suggest, in analogy to the earlier science of heat engines, that perhaps it will come to be known as Cognodynamics.

\section{Acknowledgments}

 The authors acknowledge many conversations, particularly with Eric Pulick, at University of Wisconsin, Madison (UWM) who built an evaluation framework;  Shubham Bharti. PK thanks Prof. Fred Sala at UWN for helpful conversations on the notion of fiducial points in a quasi-metric space. 
 This research has been supported in part by DARPA under other agreement HR001124203; the National Science Foundation, under NSF Grant 2041428; the Wisconsin Alumni Research Foundation.
This research was sponsored in part  by the Defense Advanced Research Projects Agency;  the content of the information does not necessarily reflect the position or the policy of the Government; and  no official endorsement should be inferred. Distribution Statement: Approved for public release; distribution is unlimited.

 % \pk{HAO are there any other sources to acknowledge? Hao: Nope. I think we are good here.} 

\printbibliography

\appendix   % <-- tells LaTeX subsequent sections are appendices
\section{Additional Experiments}

\begin{table}[htbp]
\centering
\caption{Rules used in Rule-Based experiments.}
\label{tab:rules-based exp}
\begin{tabular}{|r| l| @{\hspace{1cm}} |r |l|}
\hline
\textbf{S.No} & \textbf{Rules} & \textbf{S.No} & \textbf{Rules} \\
\hline
1  & allOfColOrd\_BRKY       & 10 & ordL1\_Nearby      \\
2  & allOfShaOrd\_qcts       & 11 & ordRevOfL1         \\
3  & ccw                     & 12 & ordRevOfL1\_Remotest\\
4  & cm\_RBKY                & 13 & quadMixed1         \\
5  & col1Ord\_BRKY           & 14 & quadNearby         \\
6  & col1OrdBuck\_BRKY0213   & 15 & sha1Ord\_qcts      \\
7  & colOrdL1\_BRKY          & 16 & sha1OrdBuck\_qcts0213 \\
8  & cw                      & 17 & shaOrdL1\_qcts     \\
9  & ordL1                   & 18 & sm\_csqt           \\
\hline
\end{tabular}
% \label{tab:rules_side_by_side}
\end{table}

% \begin{table*}[ht]
% \centering
% \caption{
%     List of rules and their descriptions.\\
% }
% \label{tab:rules-desc}

% \begin{tabular}{|l|p{0.65\textwidth}|}
% \hline
% \textbf{Rule} & \textbf{Description} \\
% \hline

\begin{longtable}{|l|p{0.60\textwidth}|}
\caption{List of rules and their descriptions.} \label{tab:rules-desc} \\
\hline
\textbf{Rule} & \textbf{Description} \\
\hline
\endfirsthead

% \hline
% \textbf{Rule} & \textbf{Description} \\
% \hline
% \endhead
allOfColOrd\_BRKY & All pieces of one color are removed before others, in a given order; Blue $\rightarrow$ Red $\rightarrow$ Black $\rightarrow$ Yellow. \\

% allOfColOrd\_KRBY & All pieces of one color before others, in order: Black $\rightarrow$ Red $\rightarrow$ Blue $\rightarrow$ Yellow. \\
% allOfShaOrd\_csqt & All pieces of one shape before others, in order: Circle $\rightarrow$ Star $\rightarrow$ Square $\rightarrow$ Triangle. \\

allOfShaOrd\_qcts & All pieces of one shape are removed before others, in given in order; Square $\rightarrow$ Circle $\rightarrow$ Triangle $\rightarrow$ Star. \\

% buckets\_2130 & Buckets filled in the order: 2 $\rightarrow$ 1 $\rightarrow$ 3 $\rightarrow$ 0 (loop). \\

ccw & Start with any bucket and ill buckets in counterclockwise order. \\

% cm\_KRBY & Match colors with buckets, any order. 0-Black; 1-Red; 2-Blue; 3-Yellow. \\

cm\_RBKY & Assign colors to specific buckets in any order; 0-Red, 1-Blue, 2-Black, 3-Yellow. \\

% cm\_RBKY\_cw\_0123 & Match colors with buckets, in given order: 0-Red $\rightarrow$ 1-Blue $\rightarrow$ 2-Black $\rightarrow$ 3-Yellow (loop). \\

col1Ord\_BRKY & Assign colors to any bucket in given order. Skip if no piece of that color; Blue $\rightarrow$ Red $\rightarrow$ Black $\rightarrow$ Yellow.  loop. \\

col1Ord\_KRBY & Assign colors to any bucket in given order. Skip if no piece of that color;  Black $\rightarrow$ Red $\rightarrow$ Blue $\rightarrow$ Yellow.  loop. \\

col1OrdBuck\_BRKY0213 & Assign colors to specific buckets in given order, : 0-Blue, 2-Red, 1-Black, 3-Yellow; loop. \\

% col1OrdBuck\_BRKY3120 & Assign colors to specific buckets in order: 3-Blue, 1-Red, 2-Black, 0-Yellow; loop. \\

colOrdL1\_BRKY & Remove colors in given order and if a color is missing take the next object in reading order of that color. Blue $\rightarrow$ Red $\rightarrow$ Black $\rightarrow$ Yellow.  loop. \\

% colOrdL1\_KBYR & Colors in order: Black $\rightarrow$ Blue $\rightarrow$ Yellow $\rightarrow$ Red. Must take next object in reading order of that color; loop. \\

cw & Start with any bucket and fill buckets in clockwise order. \\

% cw\_0123 & Start with bucket 0; fill buckets in clockwise order: 0 $\rightarrow$ 1 $\rightarrow$ 2 $\rightarrow$ 3 (loop). \\

ordL1 & Remove pieces in reading order and assign pieces to any bucket. \\

ordL1\_Nearby & Remove pieces in reading order and  assign each to the nearest bucket. \\

ordRevOfL1 & Remove pieces in reverse reading order and assign pieces to any bucket. \\

% ordRevOfL1\_Nearby & Remove pieces in reverse reading order; assign to the nearest bucket. \\

ordRevOfL1\_Remotest & Remove pieces in reverse reading order and assign to the farthest bucket. \\

quadMixed1 & Assign pieces in each quadrant to specific buckets: 0-quadrant 3, 1-quadrant 0, 2-quadrant 2, 3-quadrant 1. \\

quadNearby & Assign pieces to the nearest quadrant in any order. \\

% sha1Ord\_csqt & Assign shapes to any bucket in order: Circle $\rightarrow$ Star $\rightarrow$ Square $\rightarrow$ Triangle. Skip if none; loop. \\

sha1Ord\_qcts & Assign shapes to any bucket in given order. Skip if no piece of that shape; Square $\rightarrow$ Circle $\rightarrow$ Triangle $\rightarrow$ Star. loop. \\

sha1OrdBuck\_qcts0213 & Assign shapes to specific buckets in given order: 0-Square, 2-Circle, 1-Triangle, 3-Star; loop. \\

% sha1OrdBuck\_tqsc0213 & Assign shapes to specific buckets in order: 0-Triangle, 2-Square, 1-Star, 0-Circle; loop. \\

% shaOrdL1\_csqt & Shapes in order: Circle $\rightarrow$ Star $\rightarrow$ Square $\rightarrow$ Triangle. Must take next object in reading order of that shape; skip if none; loop. \\

shaOrdL1\_qcts & Remove shapes in given order and if a shape is missing take the next object in reading order of that shape; Square $\rightarrow$ Circle $\rightarrow$ Triangle $\rightarrow$ Star; loop. \\

sm\_csqt & Assign shapes to specific bucket in any order; 0-Circle, 1-Star, 2-Square, 3-Triangle. \\

% sm\_qcts & Match shape with bucket, any order: 0-Square; 1-Circle; 2-Triangle; 3-Star. \\

cm\_RBKY\_cw\_0123 & Assign colors to specific buckets in given order; 0-Red $\rightarrow$ 1-Blue $\rightarrow$ 2-Black $\rightarrow$ 3-Yellow (loop). \\

cw\_0123 & Start with bucket 0 and fill buckets in clockwise order: 0 $\rightarrow$ 1 $\rightarrow$ 2 $\rightarrow$ 3 (loop). \\

cm\_ordL1 & Colors are removed from board in reading order and assigned to specific buckets; 0-Blue, 1-Red, 2-Black, 3-Yellow\\

cw\_qn2 & Start with bucket 0 and fill buckets in clockwise order with pieces of nearest quadrant to that bucket.;  0-quadrant 0, 1-quadrant 1, 2-quadrant 2, 3-quadrant 3\\

\hline
\multicolumn{2}{|p{0.87\textwidth}|}{
\textbf{Note:} \newline
1) Bucket and quadrant indices range from 0 to 3. Bucket 0 and quadrant 0 correspond to the top-left of the board; 1:top-right; 2: bottom-right; 3: bottom-left. \newline
2) Reading order proceeds left-to-right, top-to-bottom. \newline
3)Unprefixed numbers (e.g., 0, 1, 2, 3) refer to bucket indices.
} \\
\hline
\end{longtable}

\begin{table}[h]
\centering
\begin{tabular}{@{}l l S[table-format=6.0]@{}}
\toprule
\textbf{Property} & \textbf{Rule} & {\textbf{$M^\star$}} \\
\midrule
Quadrant\_to\_bucket\_mapping & quadNearby & 421 \\
\rowcolor{HighlightA}Feature\_ordering & sha1OrdBuck\_qcts0213 & 426 \\
Quadrant\_to\_bucket\_mapping & quadMixed1 & 468 \\
\rowcolor{HighlightC}Feature\_ordering & shaOrdL1\_qcts & 708 \\
\rowcolor{HighlightB}Proximity & ordL1\_Nearby & 782 \\
\rowcolor{HighlightB}Proximity & ordRevOfL1\_Remotest & 795 \\
\rowcolor{HighlightA}Feature\_ordering & col1OrdBuck\_BRKY0213 & 988 \\
Reading\_order & ordL1 & 1458 \\
\rowcolor{HighlightB}Reading\_order & ordRevOfL1\_Remotest & 1628 \\
\rowcolor{HighlightB} Reading\_order & ordL1\_Nearby & 1716 \\ % <— color A example
Reading\_order & ordRevOfL1 & 1809 \\
\rowcolor{HighlightC}Feature\_ordering & colOrdL1\_BRKY & 2943 \\
Feature\_to\_bucket\_mapping & sm\_csqt & 10577 \\
Feature\_to\_bucket\_mapping & cm\_RBKY & 12278 \\
All\_pieces\_of\_feature & allOfShaOrd\_qcts & 15020 \\
All\_pieces\_of\_feature & allOfColOrd\_BRKY & 15563 \\
\rowcolor{HighlightA}Feature\_to\_bucket\_mapping & col1OrdBuck\_BRKY0213 & 22709 \\
\rowcolor{HighlightA}Feature\_to\_bucket\_mapping & sha1OrdBuck\_qcts0213 & 23040 \\
Bucket\_order\_correct & ccw & 56327 \\
Bucket\_order\_correct & cw & 63377 \\ % <— color B example
Feature\_ordering & col1Ord\_BRKY & 80000 \\
Feature\_ordering & sha1Ord\_qcts & 124342 \\
\rowcolor{HighlightC}Conditional & colOrdL1\_BRKY & 170992 \\
\rowcolor{HighlightC}Conditional & shaOrdL1\_qcts & 229772 \\
\bottomrule
\end{tabular}
\caption{Rule properties arranged in the increasing order of difficulty of $M^\star$ in FC Model. Window size for $M^\star$ is 15. The higlighted rows indicates the properties that are being combined with other properties.}
% \pk{9/6/2025 1:29:34 PM The highlighting is very hard to see.} }
\label{tab: aspects_FC}
\end{table}

\begin{table}[h]
\centering
\begin{tabular}{@{}l l S[table-format=5.0]@{}}
\toprule
\textbf{Property} & \textbf{Rule} & {\textbf{$M^\star$}} \\
\midrule
\rowcolor{HighlightA}Feature\_ordering & sha1OrdBuck\_qcts0213 & 156 \\
\rowcolor{HighlightA}Feature\_ordering & col1OrdBuck\_BRKY0213 & 526 \\
All\_pieces\_of\_feature & allOfShaOrd\_qcts & 561 \\
All\_pieces\_of\_feature & allOfColOrd\_BRKY & 611 \\
Feature\_to\_bucket\_mapping & cm\_RBKY & 662 \\
Quadrant\_to\_bucket\_mapping & quadNearby & 696 \\
Feature\_to\_bucket\_mapping & sm\_csqt & 733 \\
Quadrant\_to\_bucket\_mapping & quadMixed1 & 915 \\
Bucket\_order\_correct & ccw & 1128 \\
\rowcolor{HighlightA}Feature\_to\_bucket\_mapping & sha1OrdBuck\_qcts0213 & 1176 \\
\rowcolor{HighlightA}Feature\_to\_bucket\_mapping & col1OrdBuck\_BRKY0213 & 1212 \\
 Bucket\_order\_correct & cw & 1305 \\ % example highlight
Reading\_order & ordRevOfL1 & 1456 \\
Reading\_order & ordL1 & 1648 \\
\rowcolor{HighlightB}Reading\_order & ordRevOfL1\_Remotest & 2358 \\
\rowcolor{HighlightB}Proximity & ordRevOfL1\_Remotest & 3139 \\
\rowcolor{HighlightC}Feature\_ordering & shaOrdL1\_qcts & 3336 \\
\rowcolor{HighlightB}Proximity & ordL1\_Nearby & 3939 \\
\rowcolor{HighlightB} Reading\_order & ordL1\_Nearby & 4061 \\ % example highlight
\rowcolor{HighlightC}Feature\_ordering & colOrdL1\_BRKY & 8905 \\
Feature\_ordering & sha1Ord\_qcts & 21197 \\
 Feature\_ordering & col1Ord\_BRKY & 25980 \\ % example highlight
\rowcolor{HighlightC}Conditional & shaOrdL1\_qcts & 34115 \\
\rowcolor{HighlightC}Conditional & colOrdL1\_BRKY & 37964 \\
\bottomrule
\end{tabular}
\caption{Rule properties arranged in the increasing order of difficulty of $M^\star$ in OC Model. Window size for $M^\star$ is 15. The higlighted rows indicates the properties that are being combined with other properties.  }
\label{tab: aspects_OC}

\end{table}

\begin{table}[htb]
\centering
\begin{tikzpicture}
  % Draw the table
  \node (tab) {
    \begin{tabular}{lccc}
      \toprule
      Rule & M\_star & E\_star\_mean & E\_star\_max \\
      \midrule

      \multicolumn{4}{c}{\hrulefill\quad \textbf{Highly Learnable}\quad\hrulefill} \\
      quadNearby                &  273  &   14   &   24   \\
      quadMixed1                &  325  &   15   &   27   \\[2pt]

      \multicolumn{4}{c}{\hrulefill\quad \textbf{Moderately Learnable}\quad\hrulefill} \\
      ordL1\_Nearby             & 1189  &   62   &   73   \\
      ordL1                     & 1071  &   62   &   87   \\
      ordRevOfL1\_Remotest      & 1294  &   64   &   79   \\
      ordRevOfL1                &  984  &   75   &   88   \\[2pt]

      \multicolumn{4}{c}{\hrulefill\quad \textbf{Challenging}\quad\hrulefill} \\
      sm\_csqt                  &  9413 &  403   &  509   \\
      cm\_RBKY                  &  9527 &  426   &  440   \\
      allOfColOrd\_BRKY         &  3976 &  846   &  934   \\
      allOfShaOrd\_qcts         &  2115 &  890   & 1000   \\[2pt]

      \multicolumn{4}{c}{\hrulefill\quad \textbf{Difficult}\quad\hrulefill} \\
      col1OrdBuck\_BRKY0213     & 28112 & 1865   & 2161   \\
      sha1OrdBuck\_qcts0213     & 30993 & 2017   & 2321   \\
      ccw                       & 28265 & 3269   & 3372   \\
      cw                        & 32638 & 3592   & 3869   \\[2pt]

      \multicolumn{4}{c}{\hrulefill\quad \textbf{Very Difficult}\quad\hrulefill} \\
      sha1Ord\_qcts             &  2377 & 14145  & 13632  \\
      col1Ord\_BRKY             &  3382 & 17353  & 17572  \\
      colOrdL1\_BRKY            & 58034 & 30268  & 28992  \\
      shaOrdL1\_qcts            & 35089 & 33004  & 32998  \\

      \bottomrule
    \end{tabular}
  };
  % Arrow indicating increasing difficulty
  \draw[line width=1.5pt,
      -{Stealth[length=12pt,width=12pt]}]  ([xshift=-7.2cm]tab.north) -- ([xshift=-7.2cm]tab.south)
        node[midway,left,xshift=-3mm,rotate=90]{\textbf{Increasing Difficulty}};
\end{tikzpicture}
\caption{Rules are ordered from easiest (top) to hardest (bottom) for the FC transformer model. The ordering is based on majority ranking across the three performance metrics, where lower values denote better performance.The values shown ($E^\star_{\text{mean}}$, $E^\star_{\text{max}}$, $M^\star$) are medians taken across five independent runs. It can be shown that the range of even such a small set provides about a 94\% confidence interval for the estimate of the median.}
% \christo{modified the column names to E\_star\_mean and E\_star\_max, also removed Prefix FC\_. Also modified the caption}.\pk{9/6/2025 1:26:42 PM OK.. Thanks} }
\label{tab:FC_difficulty}
\end{table}

\begin{table}[htb]
\centering
\begin{tikzpicture}
  \node (tab) {
    \begin{tabular}{lccc}
      \toprule
      Rule & M\_star & E\_star\_mean & E\_star\_max \\
      \midrule
      \multicolumn{4}{c}{\hrulefill\quad \textbf{ Learnable}\quad\hrulefill} \\
      
      cm\_RBKY                  & 565      & 17     & 28     \\
      sm\_csqt                  & 683      & 20     & 26     \\
      allOfShaOrd\_qcts         & 548      & 30     & 44     \\
      quadNearby                & 667      & 33     & 42     \\
      allOfColOrd\_BRKY         & 476      & 37     & 47     \\
      quadMixed1                & 829      & 36     & 50     \\
      ccw                       & 612      & 60     & 85     \\
      cw                        & 738      & 53     & 68     \\
 % \multicolumn{4}{c}{\hrulefill\quad \textbf{Learnable}\quad\hrulefill} \\

      ordL1                     & 689      & 74     & 99     \\
      ordRevOfL1                & 807      & 87     & 122    \\
      \multicolumn{4}{c}{\hrulefill\quad \textbf{Challenging / Conditional}\quad\hrulefill} \\
      ordRevOfL1\_Remotest      & 2129     & 136    & 233    \\
      ordL1\_Nearby             & 2478     & 149    & 183    \\
      col1OrdBuck\_BRKY0213     & 1816     & 554    & 568    \\
      sha1OrdBuck\_qcts0213     & 2729     & 395    & 556    \\
      \multicolumn{4}{c}{\hrulefill\quad \textbf{Difficult }\quad\hrulefill} \\
      sha1Ord\_qcts             & 4530     & 3294   & 4113   \\
      col1Ord\_BRKY             & 5058     & 2114   & 2891   \\
      shaOrdL1\_qcts            & 13057    & 3205   & 3262   \\
      colOrdL1\_BRKY            & 13110    & 3232   & 3675   \\
      \bottomrule
    \end{tabular}
  };
  \draw[line width=1.5pt,
      -{Stealth[length=12pt,width=12pt]}] ([xshift=-7.2cm,yshift=0cm]tab.north) -- ([xshift=-7.2cm,yshift=0.1cm]tab.south) 
      node[midway,left,xshift=-3mm,rotate=90,black]{\textbf{Increasing Difficulty}};
\end{tikzpicture}
\caption{Rules are ordered from easiest (top) to hardest (bottom) for the OC transformer model. The ordering is based on majority ranking across the three performance metrics, where lower values denote better performance.  The values shown ($E^\star_{\text{mean}}$, $E^\star_{\text{max}}$, $M^\star$) are medians taken across five independent runs. It can be shown that the range of even such a small set provides about a 94\% confidence interval for the estimate of the median.}
% \christo{modified the column names to E\_star\_mean and E\_star\_max, also removed OC\_ from the column names. Also modified the caption}}
%\rc{Perhaps add to caption: The entry in each cell is the median value of that metric, for a set of five independent runs. It can be shown that the range of even such a small set provides a 90\% confidence interval for the estimate of the median. check whether it is 90\$ or even better at 95\%}}
\label{tab:OC_difficulty}
\end{table}

\begin{table}[htb]
    \centering
    \caption{Transfer Experiments (A)Sequential Transfer Experiments.
    The combination of rules, A+B, is represented as A\_B in the table. E.g. cm\_ordL1 represents the combination of cm\_RBKY and ordL1 }
    \label{tab:sequential_transfer}
    
    \footnotesize
    % ===== CMPND_2 =====
    \renewcommand{\arraystretch}{1.3} % (optional) for row height
    
    \begin{tabular}{|c|p{6.3 cm}||c|p{6.3 cm}|}
    \hline
    \multicolumn{4}{|c|}{\textbf{CMPND\_2}} \\ % <-- THIS IS NOW INSIDE tabular!
    \hline
    \multicolumn{2}{|c||}{\textbf{List 1}} & \multicolumn{2}{|c|}{\textbf{List 2}} \\
    \hline
    \textbf{S.No} & \textbf{Rules} & \textbf{S.No} & \textbf{Rules} \\
    \hline
    1 & quadNearby & 1 & cm\_RBKY \\
    2 & quadNearby:ordL1 & 2 & cm\_RBKY:cw\_0123 \\
    3 & quadNearby:ordL1:ordL1\_Nearby & 3 & cm\_RBKY:cw\_0123:cm\_RBKY\_cw\_0123 \\
    4 & quadNearby:ordL1:ordL1\_Nearby: \mbox{cm\_RBKY\_cw\_0123} & 4 & 
    \mbox{cm\_RBKY:cw\_0123:cm\_RBKY\_cw\_0123:} \mbox{ordL1\_Nearby} \\

    \hline\hline
    \multicolumn{2}{|c||}{\textbf{List 3}} & \multicolumn{2}{c|}{\textbf{List 4}} \\
    \hline
    \textbf{S.No} & \textbf{Rules} & \textbf{S.No} & \textbf{Rules} \\
    \hline
    1 & quadNearby & 1 & cm\_RBKY \\
    2 & quadNearby:ordL1 & 2 & cm\_RBKY:cw\_0123 \\
    3 & quadNearby:ordL1:cm\_RBKY\_cw\_0123 & 3 & cm\_RBKY:cw\_0123:ordL1\_Nearby \\
    4 & \mbox{quadNearby:ordL1:cm\_RBKY\_cw\_0123:} \mbox{ordL1\_Nearby} & 4 & \mbox{cm\_RBKY:cw\_0123:ordL1\_Nearby:} \mbox{cm\_RBKY\_cw\_0123} \\
    \hline

    \hline\hline
    
    % ===== CMPND_3 =====
    \multicolumn{4}{|c|}{\textbf{CMPND\_3}} \\
    \hline
    \multicolumn{2}{|c||}{\textbf{List 1}} & \multicolumn{2}{c|}{\textbf{List 2}} \\
    \hline
    \textbf{S.No} & \textbf{Rules} & \textbf{S.No} & \textbf{Rules} \\
    \hline
    1 & cm\_RBKY & 1 & cm\_RBKY \\
    2 & cm\_RBKY:ordL1 & 2 & cm\_RBKY:ordL1 \\
    3 & cm\_RBKY:ordL1:cm\_ordL1 & 3 & cm\_RBKY:ordL1:cw\_qn2\\
    4 & cm\_RBKY:ordL1:cm\_ordL1:cw\_qn2& 4 & cm\_RBKY:ordL1:cw\_qn2:cm\_ordL1 \\
    \hline\hline

    \multicolumn{2}{|c||}{\textbf{List 3}} & \multicolumn{2}{c|}{\textbf{List 4}} \\
    \hline
    \textbf{S.No} & \textbf{Rules} & \textbf{S.No} & \textbf{Rules} \\
    \hline
    1 & cw & 1 & cw \\
    2 & cw:quadNearby & 2 & cw:quadNearby \\
    3 & cw:quadNearby:cm\_ordL1 & 3 & cw:quadNearby:cw\_qn2 \\
    4 & cw:quadNearby:cm\_ordL1:cw\_qn2 & 4 & cw:quadNearby:cw\_qn2:cm\_ordL1 \\
    \hline
    \end{tabular}
\end{table}

% \label{tab:experiment7}

\begin{table}[htb] % \label{tab:partial_transfer}
    \centering
    \caption{Transfer Experiments (B) Partial Transfer Experiments 
    The combination of rules, A+B, is represented as A\_B in the table. Eg. cm\_ordL1 represents the combination of cm\_RBKY and ordL1  
    % \pk{9/6/2025 1:27:28 PM OK } 
    }
    \label{tab:partial_transfer}
    
    \footnotesize
    
    % \footnotesize
    % % List 1 and List 3 side by side
    % \begin{center}
    %     \textbf{CMPND\_4}
    % \end{center}
    % \begin{tabular}{|c|p{6.5 cm}||c|p{6.5 cm}|}
    % ===== CMPND_4 =====
    % \multicolumn{4}{|c|}{\textbf{CMPND\_4}} \\
    
    \renewcommand{\arraystretch}{1.3} % (optional) for row height
    
    \begin{tabular}{|c|p{6.3 cm}||c|p{6.3 cm}|}
    \hline
    \multicolumn{4}{|c|}{\textbf{CMPND\_4}} \\ % <-- THIS IS NOW INSIDE tabular!
    \hline
    \hline
    \multicolumn{2}{|c||}{\textbf{List 1}} & \multicolumn{2}{c|}{\textbf{List 2}} \\
    \hline
    \textbf{S.No} & \textbf{Rules} & \textbf{S.No} & \textbf{Rules} \\
    \hline
    1 & quadNearby & 1 & cm\_RBKY \\
    2 & quadNearby:cw\_0123 & 2 & cm\_RBKY:ordL1 \\
    3 & quadNearby:cw\_0123:ordL1\_Nearby & 3 & cm\_RBKY:ordL1:cm\_RBKY\_cw\_0123 \\
    4 & \mbox{quadNearby:cw\_0123:ordL1\_Nearby:} \mbox{cm\_RBKY\_cw\_0123} & 4 & \mbox{cm\_RBKY:ordL1:cm\_RBKY\_cw\_0123:} \mbox{ordL1\_Nearby} \\
    \hline

    \hline
    \multicolumn{2}{|c||}{\textbf{List 3}} & \multicolumn{2}{c|}{\textbf{List 4}} \\
    \hline
    \textbf{S.No} & \textbf{Rules} & \textbf{S.No} & \textbf{Rules} \\
    \hline
    1 & cw\_0123 & 1 & ordL1 \\
    2 & cw\_0123:quadNearby & 2 & ordL1:cm\_RBKY \\
    3 & cw\_0123:quadNearby:cm\_RBKY\_cw\_0123 & 3 & ordL1:cm\_RBKY:ordL1\_Nearby \\
    4 & \mbox{cw\_0123:quadNearby:cm\_RBKY\_cw\_0123:} \mbox{ordL1\_Nearby} & 4 & \mbox{ordL1:cm\_RBKY:ordL1\_Nearby:} \mbox{cm\_RBKY\_cw\_0123} \\
    \hline
    % \end{tabular}
    % \label{tab:rules_trial_lists}
    % \end{table*}
    % \footnotesize
    
    % \begin{table*}
    % \footnotesize
    % \begin{center}
    %     \textbf{CMPND\_5}
    % \end{center}
    % \begin{tabular}{|c|p{6.5 cm}||c|p{6.5 cm}|}
    \multicolumn{4}{|c|}{\textbf{CMPND\_5}} \\
    \hline
    \multicolumn{2}{|c||}{\textbf{List 1}} & \multicolumn{2}{c|}{\textbf{List 2}} \\
    \hline
    \textbf{S.No} & \textbf{Rules} & \textbf{S.No} & \textbf{Rules} \\
    \hline
    1 & cm\_RBKY & 1 & cw \\
    2 & cm\_RBKY:cw & 2 & cw:cm\_RBKY \\
    3 & cm\_RBKY:cw:cm\_ordL1 & 3 & cw:cm\_RBKY:cw\_qn2\\
    4 & cm\_RBKY:cw:cm\_ordL1:cw\_qn2& 4 & cw:cm\_RBKY:cw\_qn2:cm\_ordL1 \\
    \hline\hline
    % \end{tabular}
    
    % \vspace{2em}
    
    % \begin{tabular}{|c|p{6.5 cm}|c|p{6.5 cm}|}
    % \hline
    \multicolumn{2}{|c||}{\textbf{List 3}} & \multicolumn{2}{c|}{\textbf{List 4}} \\
    \hline
    \textbf{S.No} & \textbf{Rules} & \textbf{S.No} & \textbf{Rules} \\
    \hline
    1 & ordL1 & 1 & quadNearby \\
    2 & ordL1:quadNearby & 2 & quadNearby:ordL1 \\
    3 & ordL1:quadNearby:cm\_ordL1 & 3 & quadNearby:ordL1:cw\_qn2 \\
    4 & ordL1:quadNearby:cm\_ordL1:cw\_qn2 & 4 & quadNearby:ordL1:cw\_qn2:cm\_ordL1 \\
    \hline
    \end{tabular}
\end{table}

% ---------- Page 1 (subfigures a–b) ----------
\begin{figure*}[htb]
    \centering

    \begin{subfigure}{0.90\linewidth}
        \centering
        \includegraphics[width=\linewidth]{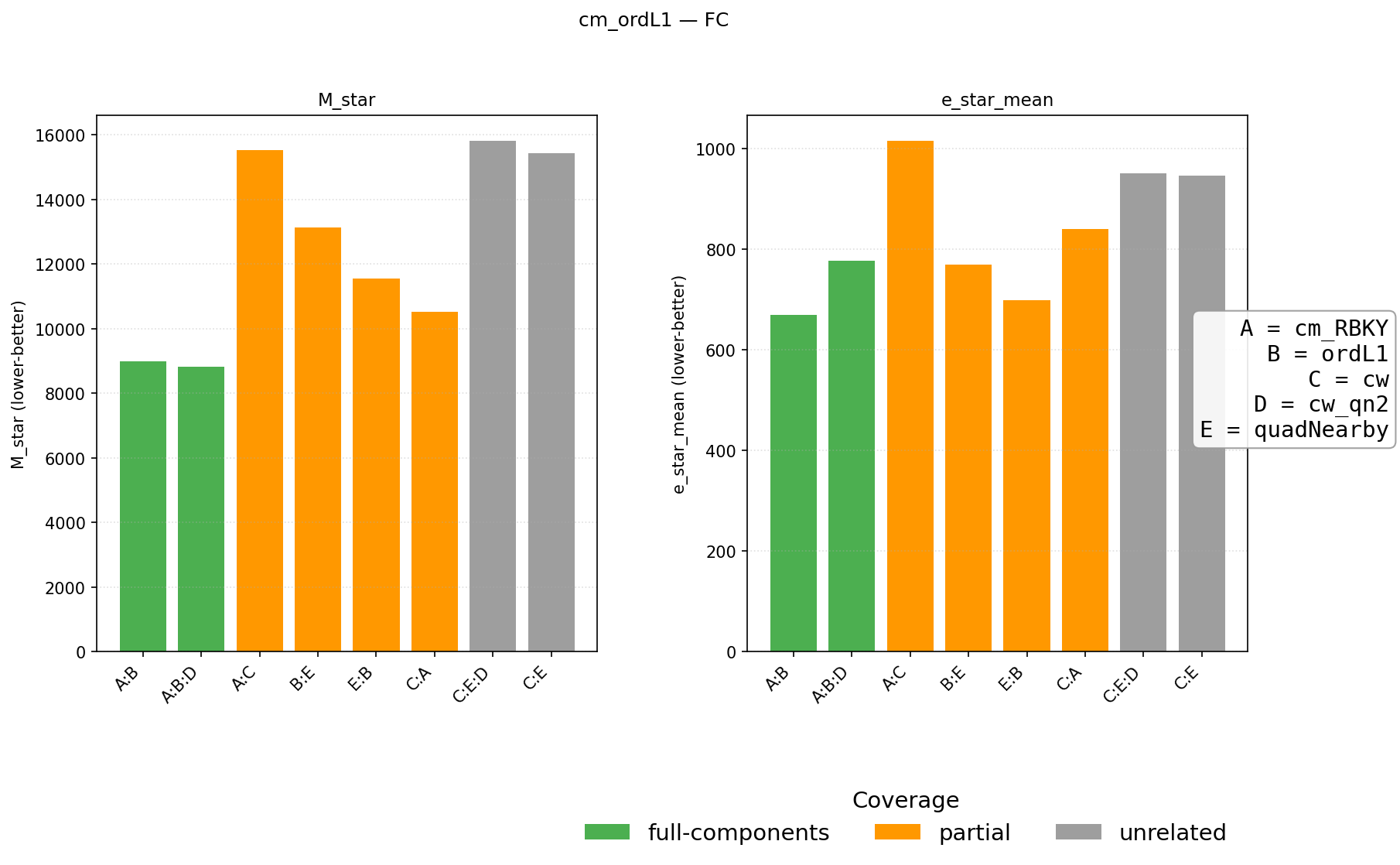}
        \caption{Performance metrics for rule \texttt{cm+ordL1} in the FC model.}
        \label{fig:fc_cm_ordL1}
    \end{subfigure}

    \vspace{1em}
    \begin{subfigure}{0.90\linewidth}
        \centering
        \includegraphics[width=\linewidth]{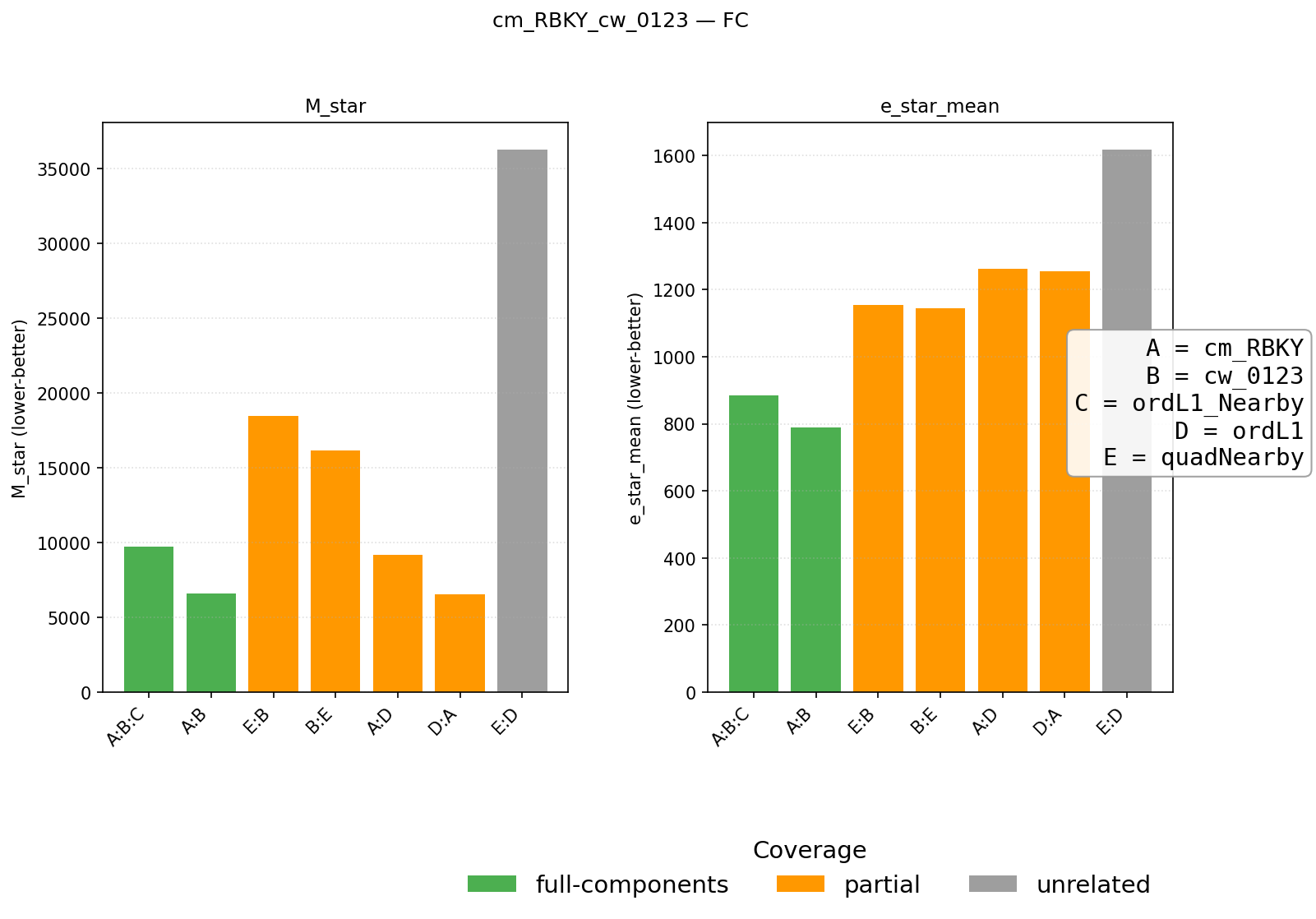}
        \caption{Performance metrics for rule \texttt{cm\_RBKY+cw\_0213} in the FC model.}
        \label{fig:fc_cm_cw}
    \end{subfigure}

    \caption{Performance metrics for compound rules in FC (top rows) and OC (bottom rows) models. Each subfigure shows results for one compound rule under different transfer curricula.}
\end{figure*}

% ---------- Page 2 (subfigures c–d) ----------
\begin{figure*}[htb]\ContinuedFloat
    \centering
    \setcounter{subfigure}{2} % continue lettering from (b) → start at (c)

    \begin{subfigure}{0.90\linewidth}
        \centering
        \includegraphics[width=\linewidth]{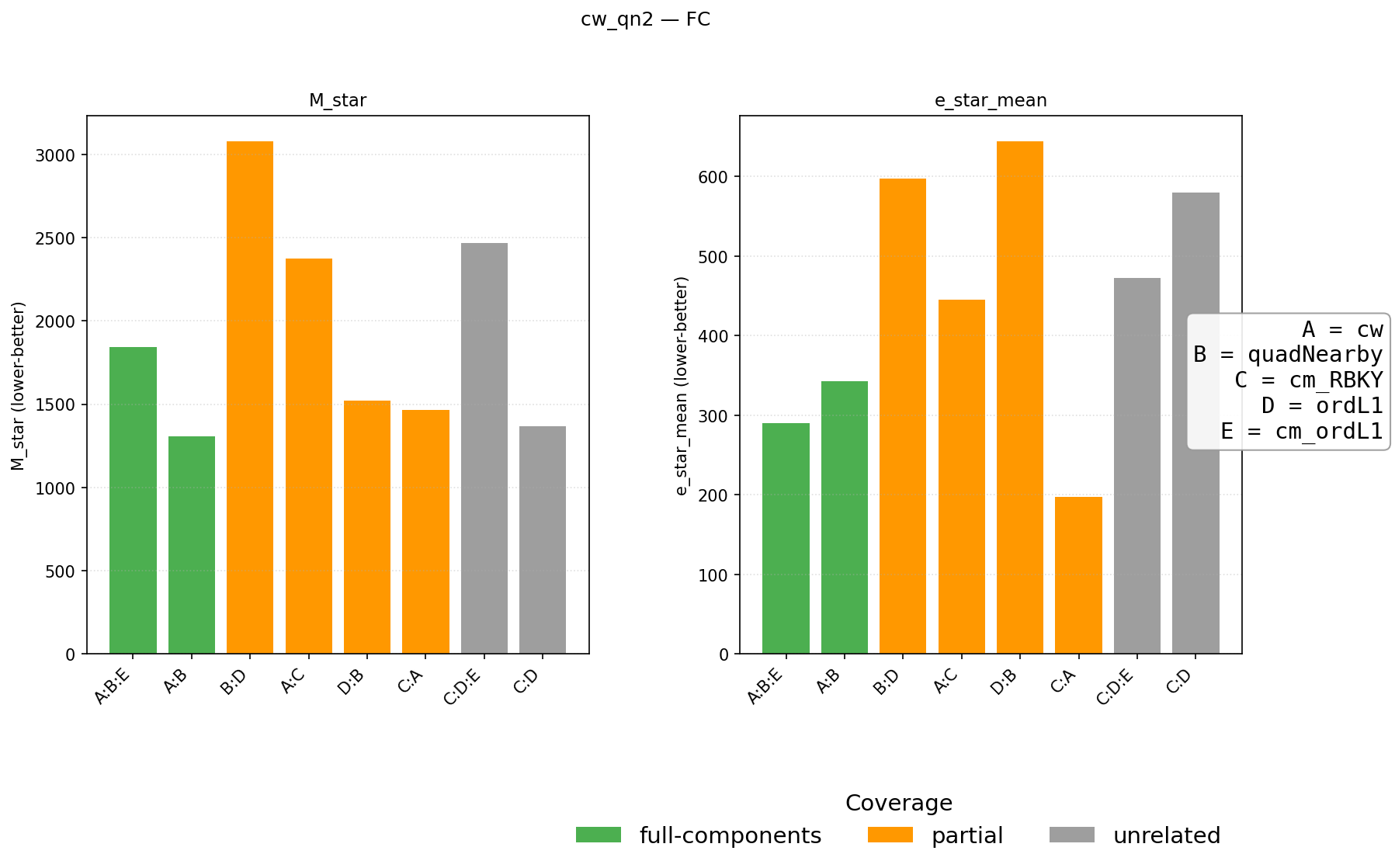}
        \caption{Performance metrics for rule \texttt{cw+quadNearby} in the FC model.}
        \label{fig:fc_cwqn2}
    \end{subfigure}

    \vspace{1em}
    \begin{subfigure}{0.90\linewidth}
        \centering
        \includegraphics[width=\linewidth]{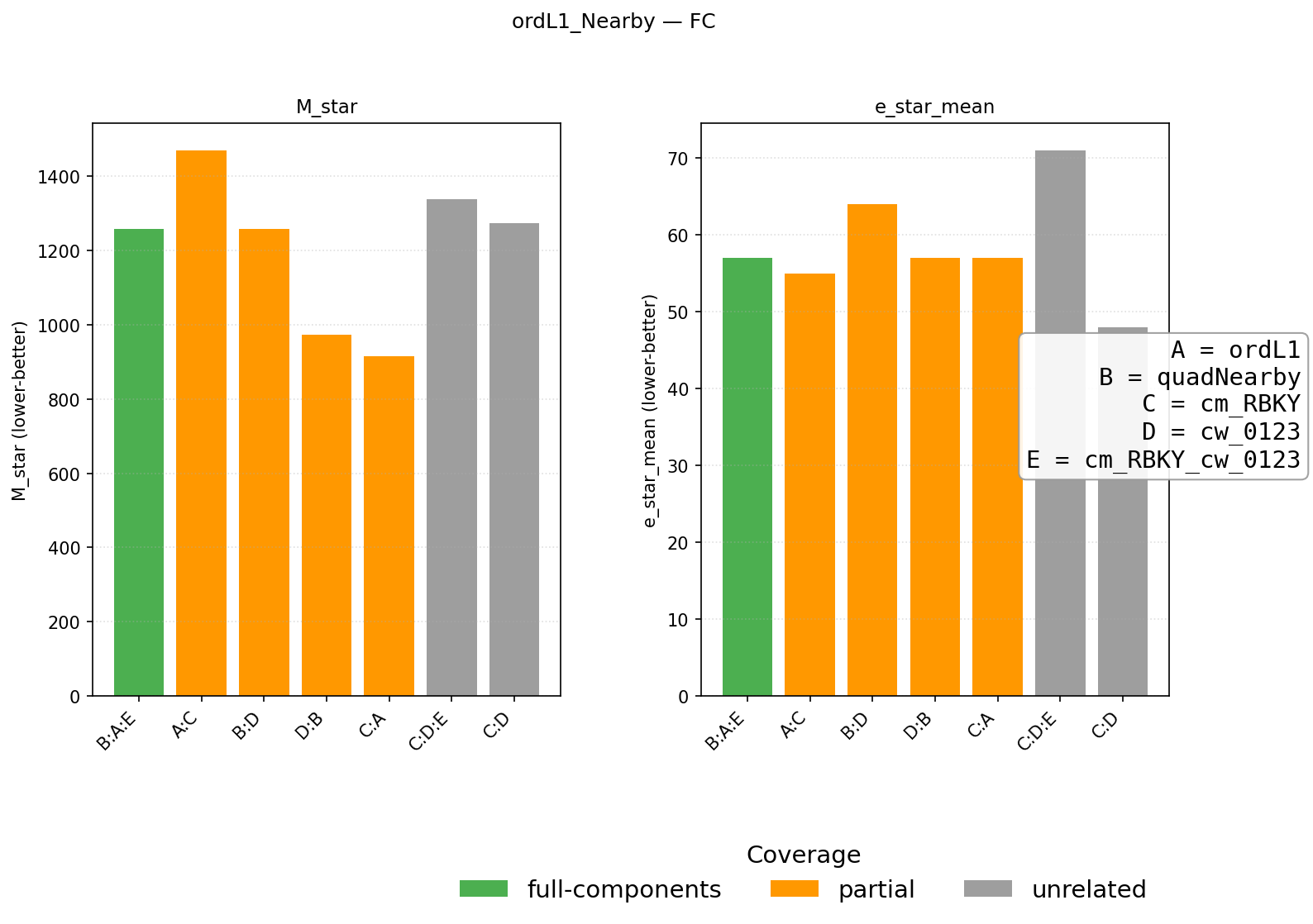}
        \caption{Performance metrics for rule \texttt{ordL1+quadNearby} in the FC model.}
        \label{fig:fc_ordL1Nearby}
    \end{subfigure}

    \caption[]{Performance metrics for compound rules in FC/OC models (continued).}
\end{figure*}

% ---------- Page 3 (subfigures e–f) ----------
\begin{figure*}[htb]\ContinuedFloat
    \centering
    \setcounter{subfigure}{4} % continue lettering from (d) → start at (e)

    \begin{subfigure}{0.90\linewidth}
        \centering
        \includegraphics[width=\linewidth]{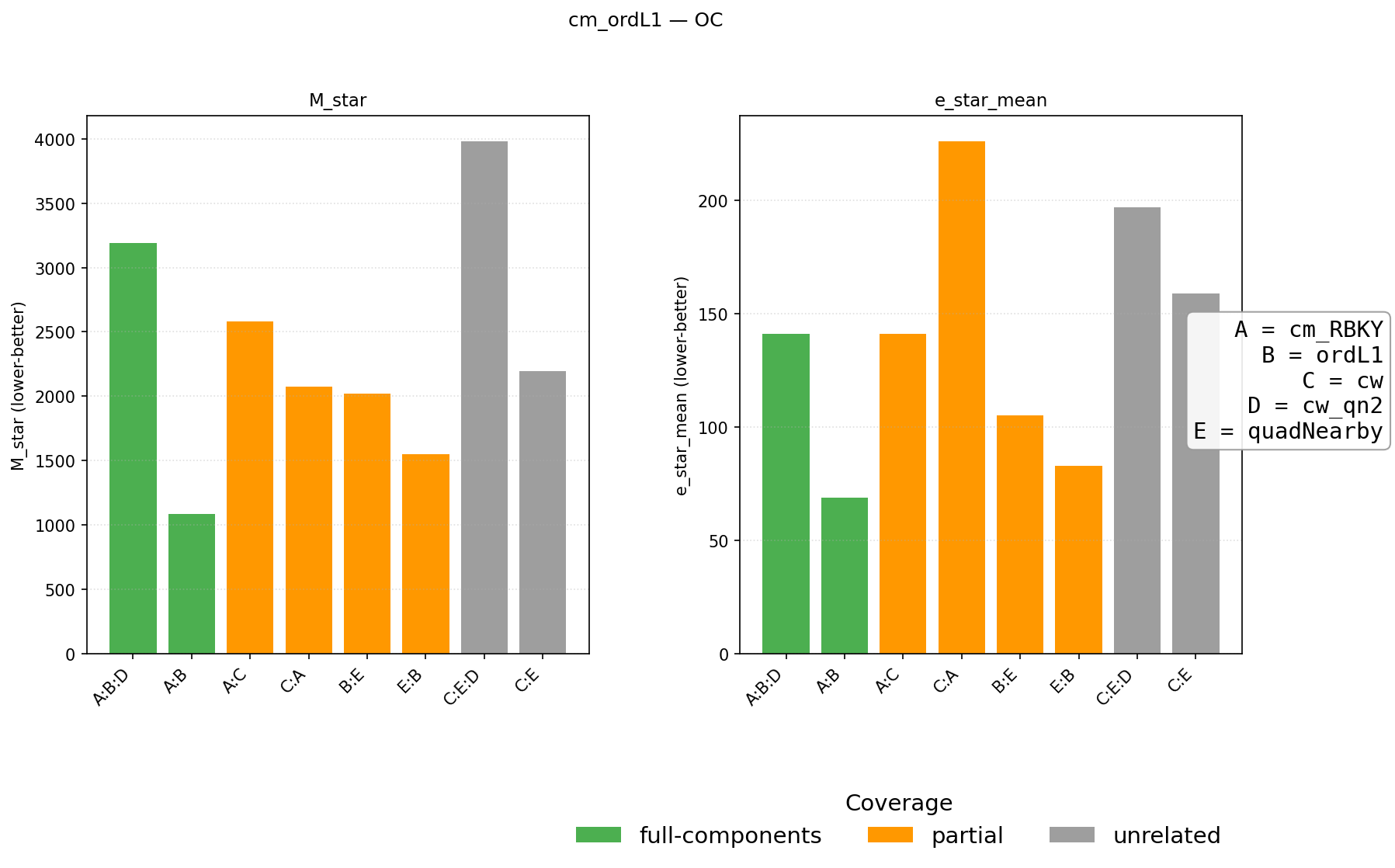}
        \caption{Performance metrics for rule \texttt{cm+ordL1} in the OC model.}
        \label{fig:oc_cm_ordL1}
    \end{subfigure}

    \vspace{1em}
    \begin{subfigure}{0.90\linewidth}
        \centering
        \includegraphics[width=\linewidth]{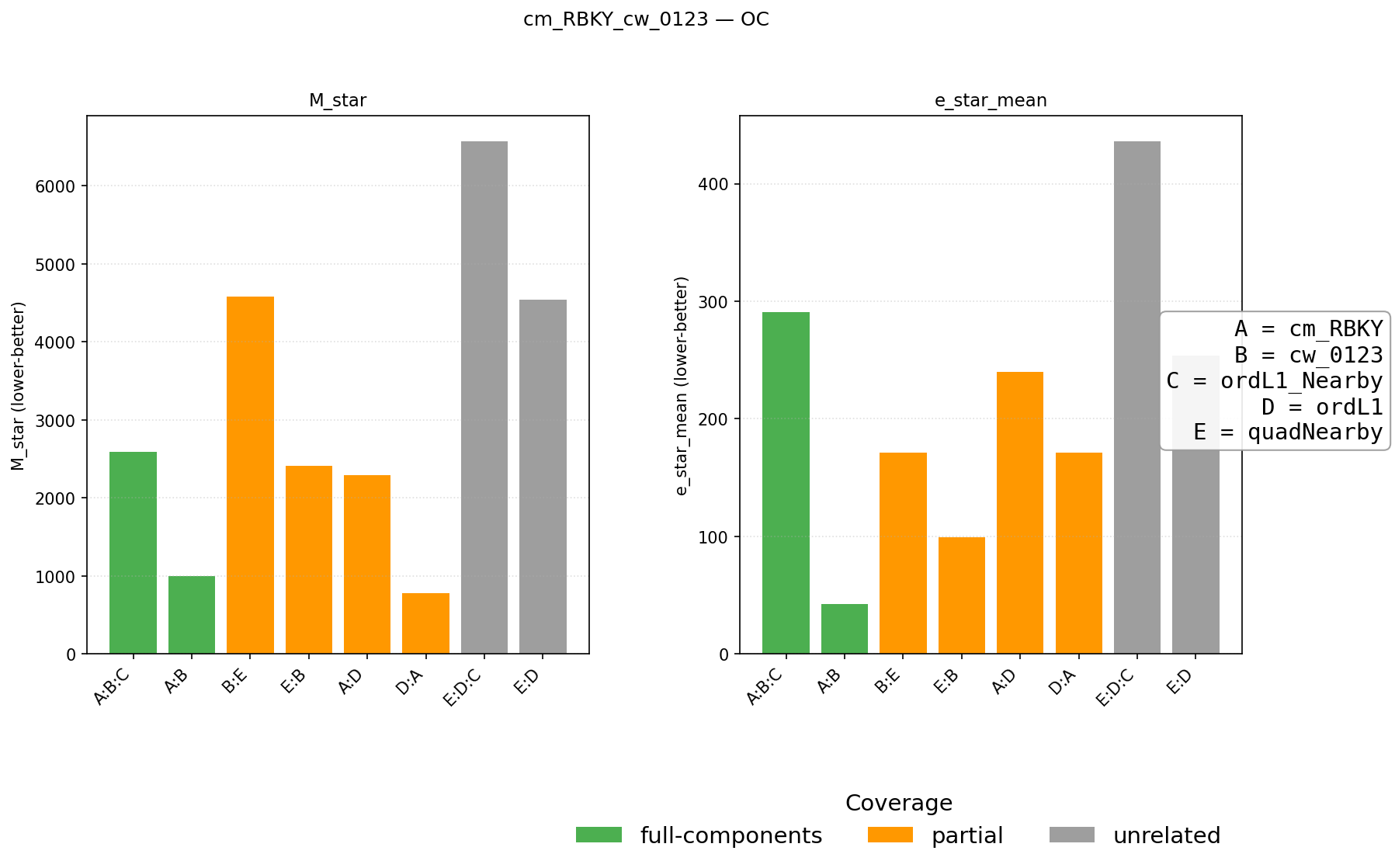}
        \caption{Performance metrics for rule \texttt{cm\_RBKY+cw\_0213} in the OC model.}
        \label{fig:oc_cm_cw}
    \end{subfigure}

    \caption[]{Performance metrics for compound rules in FC/OC models (continued).}
\end{figure*}

% ---------- Page 4 (subfigures g–h) ----------
\begin{figure*}[htb]\ContinuedFloat
    \centering
    \setcounter{subfigure}{6} % continue lettering from (f) → start at (g)

    \begin{subfigure}{0.90\linewidth}
        \centering
        \includegraphics[width=\linewidth]{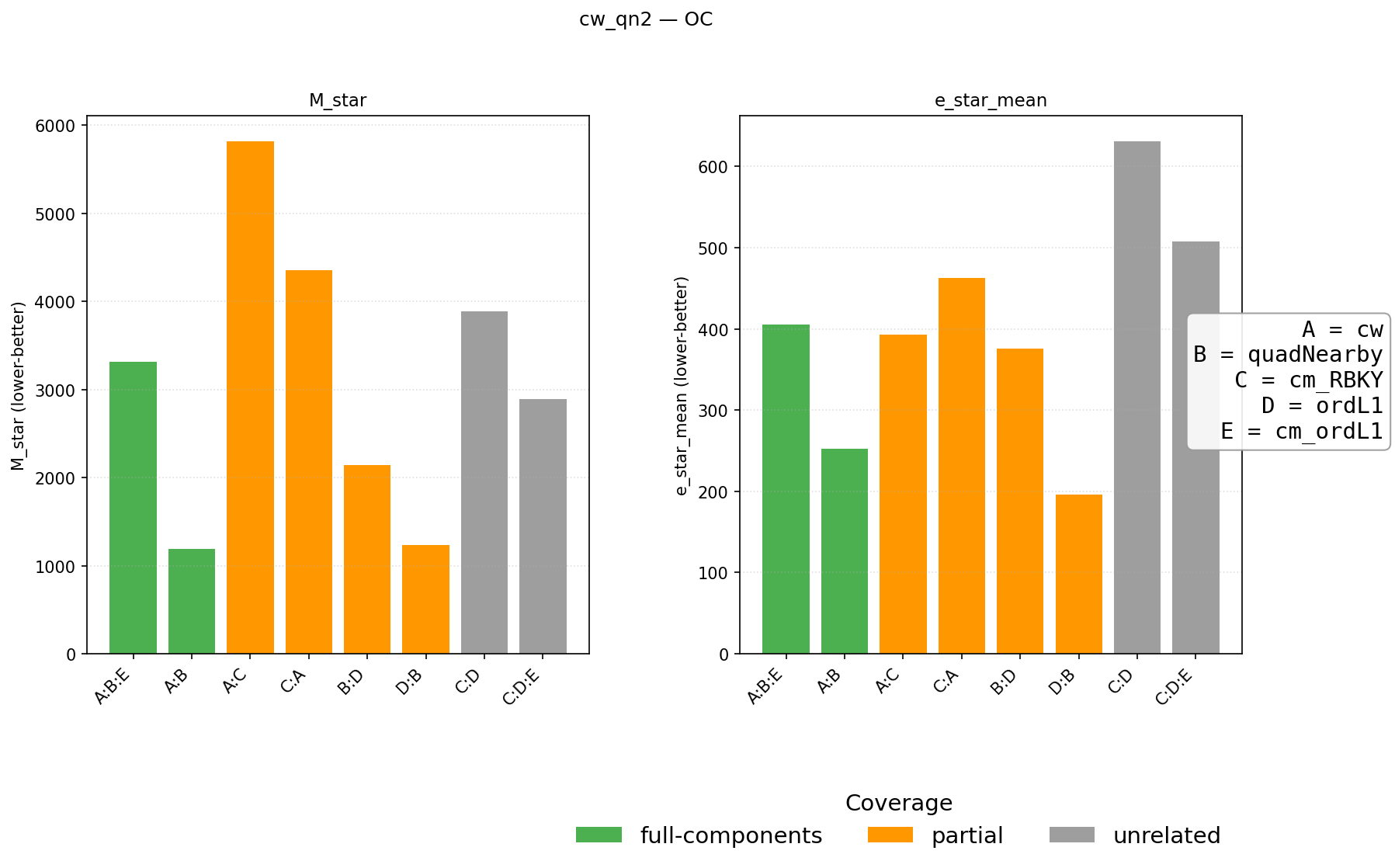}
        \caption{Performance metrics for rule \texttt{cw+quadNearby} in the OC model.}
        \label{fig:oc_cwqn2}
    \end{subfigure}

    \vspace{1em}
    \begin{subfigure}{0.90\linewidth}
        \centering
        \includegraphics[width=\linewidth]{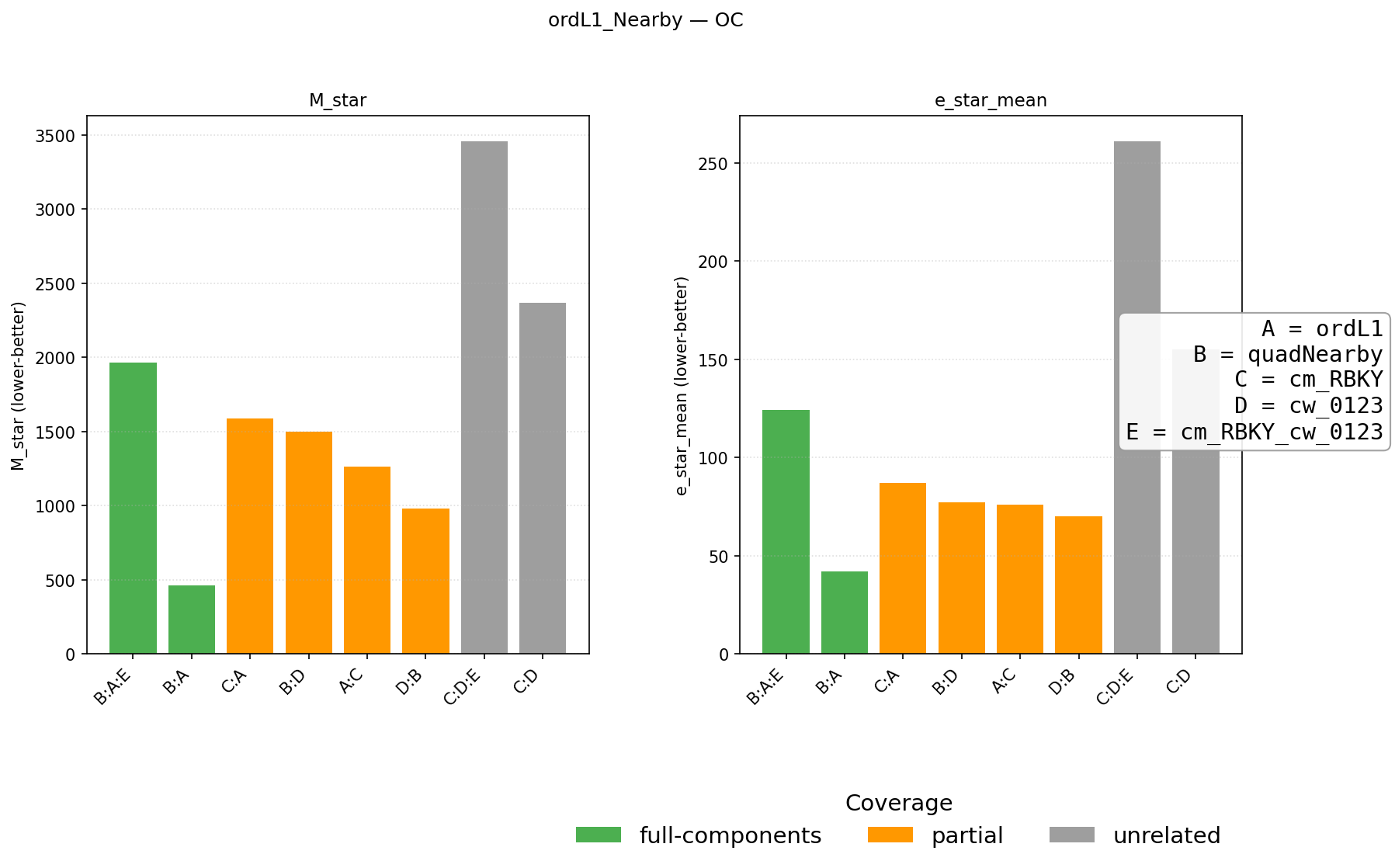}
        \caption{Performance metrics for rule \texttt{ordL1+quadNearby} in the OC model.}
        \label{fig:oc_ordL1Nearby}
    \end{subfigure}

    \caption[]{Performance metrics for compound rules in FC/OC models (continued).}
    \label{fig:transfer_all}

\end{figure*}

\end{document}